\makeatletter
\let\submissionoutputpage\@outputpage
\makeatother
\RequirePackage{fix-cm}
\documentclass{ws-us}
\usepackage[T1]{fontenc}
\usepackage{tgtermes,tgcursor}
\usepackage{mathtools,booktabs,algorithm,algpseudocode,etoolbox}
\usepackage[sort,compress,super]{cite}
\usepackage{url,balance}
\usepackage[hidelinks]{hyperref}
\hypersetup{pdftitle={Recursive Bayesian Monitoring of SLAM Fault for LiDAR-Based UAV Navigation},
  pdfauthor={Mingqiao Han, Qigeng Duan, Xuchen Liu, Jinqiang Cui, Xi Chen, and Ben M. Chen}}

\makeatletter
\let\@outputpage\submissionoutputpage
\let\trimmarks\relax
\patchcmd{\@maketitle}{\vspace*{70pt}}{\vspace*{10pt}}{}{}
\long\def\@makecaption#1#2{%
  \ifx\FigName\@captype\vskip\abovecaptionskip\fi
  \begingroup
    \ifx\FigName\@captype\figurecaptionfont\else\tablecaptionfont\fi
    \setbox\@tempboxa\hbox{#1.\quad #2}%
    \ifdim\wd\@tempboxa>\hsize\noindent #1.\quad #2\par
    \else\hbox to\hsize{\hfil\box\@tempboxa\hfil}\fi
  \endgroup
  \ifx\FigName\@captype\else\vskip5pt\fi}
\makeatother
\newcommand{\firstdisplay}[2]{#1~\ref{#2}\label{mention:#2}}
\newsavebox{\portraitbox}
\newcount\portraitlines
\newenvironment{biography}[2]{%
  \par\addvspace{8pt}\noindent
  \begin{minipage}[t]{\linewidth}\small
    \let\author\textbf
    \sbox{\portraitbox}{#1}%
    \portraitlines=\numexpr
      (\dimexpr\ht\portraitbox+\dp\portraitbox+0.35\baselineskip\relax)
      /(\baselineskip)+1\relax
    \hangindent=\dimexpr\wd\portraitbox+7pt\relax \hangafter=-\portraitlines
    \noindent\llap{\raisebox{\dimexpr\ht\strutbox-\ht\portraitbox\relax}[0pt][0pt]{\usebox{\portraitbox}}\hspace{7pt}}#2\par
  \end{minipage}\par}{}
\renewcommand{\titlefont}{\centering\fontsize{14}{16}\selectfont\bfseries\boldmath}
\usepackage{microtype}
\usepackage{stfloats}
\makeatletter
\renewcommand\section{\@startsection{section}{1}{\z@}{-14pt plus -2pt minus -2pt}{8pt}{\sectionfont}}
\renewcommand\subsection{\@startsection{subsection}{2}{\z@}{-12pt plus -2pt minus -2pt}{6pt}{\subsectionfont}}
\renewcommand\subsubsection{\@startsection{subsubsection}{3}{\z@}{-10pt plus -2pt minus -2pt}{5pt}{\subsubsectionfont}}
\makeatother
\renewcommand{\bibfont}{\fontsize{9}{10.5}\selectfont}
\begin{document}
\markboth{M. Han et al.}{Manuscript}
\twocolumn[{%
  \title{Recursive Bayesian Monitoring of SLAM Failures for LiDAR-Based UAV Navigation}
  \author{Mingqiao Han$^{a}$, Qigeng Duan$^{a}$, Xuchen Liu$^{b,*}$, Jinqiang Cui$^{b}$, Xi Chen$^{a}$, and Ben M. Chen$^{a,*}$}
  \address{$^{a}$Department of Mechanical and Automation Engineering, The Chinese University of Hong Kong, Shatin, N.T., Hong Kong, China}
  \address{$^{b}$Pengcheng Laboratory, Shenzhen, Guangdong, China}
  \address{$^{*}$Corresponding authors: Xuchen Liu (\href{mailto:liuxch@pcl.ac.cn}{liuxch@pcl.ac.cn}) and Ben M. Chen (\href{mailto:bmchen@cuhk.edu.hk}{bmchen@cuhk.edu.hk})}
  \maketitle
  \begin{abstract}
SLAM failures remain difficult to avoid in autonomous robotic systems, and undetected failures can propagate to unsafe planning, control, and mission-level decisions. Existing monitoring strategies often depend on prior maps, estimator-internal variables, or redundant sensing, which limits their applicability to deployed and resource-constrained robots. This paper presents a scene-geometry consistency monitor for detecting geometry-manifested SLAM failures using only the geometric output of the SLAM backend. The targeted failures include cumulative drift-induced failures, abrupt misregistration failures, and backend map corruption when they become observable as inconsistency in reconstructed scene geometry. The proposed method represents revisited local regions with recursive latent voxel geometry and evaluates each current observation against a trusted continuation explanation and a weak non-propagated reference explanation. The resulting voxel-wise evidence contrasts are aggregated into a fixed-boundary monitoring signal for online failure declaration and downstream safety response. Experiments on self-collected sequences, GEODE benchmark sequences, multiple robotic platforms, and different SLAM backends show that the monitor responds to failures when they are manifested in backend scene geometry. It achieves 4/4 valid post-onset declarations with no pre-onset alarms on four self-collected sequences, detects geometry-manifested failures in all five corrupted GEODE runs while remaining alarm-free in five geometrically consistent runs, adds only 0.1--0.2\% absolute CPU load, and supports a closed-loop UAV detect--isolate--recover workflow.
  \end{abstract}
}]

\section{Introduction}
\label{sec:introduction}

\noindent\textbf{Online monitoring of SLAM integrity is essential for safe autonomous flight.}
During autonomous flight, an unmanned aerial vehicle (UAV) relies on LiDAR--inertial odometry and reconstructed geometry for navigation and obstacle avoidance~\cite{fastlio2,faster_motion}. Each estimated pose determines how LiDAR returns enter the map, directly affecting the geometric information supplied to downstream planning and control. As illustrated in \firstdisplay{Figure}{introduction}, a pose error on a return pass can displace observations of a previously mapped surface. An online monitor must identify such geometric inconsistency when the estimator continues publishing odometry despite loss of registration trust.

\noindent\textbf{Modern SLAM remains vulnerable to cumulative drift and abrupt failure.}
Severe vibration or sudden loss of geometric constraints can abruptly displace registered surfaces. Cumulative drift develops differently: small errors may remain inconspicuous between neighboring scans but emerge when a later revisit places an existing structure at an incompatible location. Detecting both failures therefore requires geometric evidence accumulated across observations. However, even nominal revisits change voxel statistics through viewpoint variation, occlusion, and nonuniform sampling. A shift in centroid alone cannot reliably distinguish failure from benign geometric variation.

\noindent\textbf{Existing monitoring strategies face structural limitations.}
Estimator-based integrity methods propagate fault hypotheses to assess positioning uncertainty~\cite{osa1,osa2,osa3}. Registration indicators evaluate geometric constraints on motion~\cite{related_match1,related_match2}, whereas prior-map approaches verify alignment against an external geometric reference~\cite{related_prior2,prior}. These strategies depend on estimator-specific information, local registration conditions, or maps established before deployment. In unfamiliar environments, a UAV instead constructs its map online~\cite{voxelmap}. Registered point clouds offer a common monitoring substrate across SLAM backends, but their use as an integrity reference introduces a critical question: which observations should be allowed to update the reference?

\begin{figure}[t]
\centering
\includegraphics[width=1\linewidth]{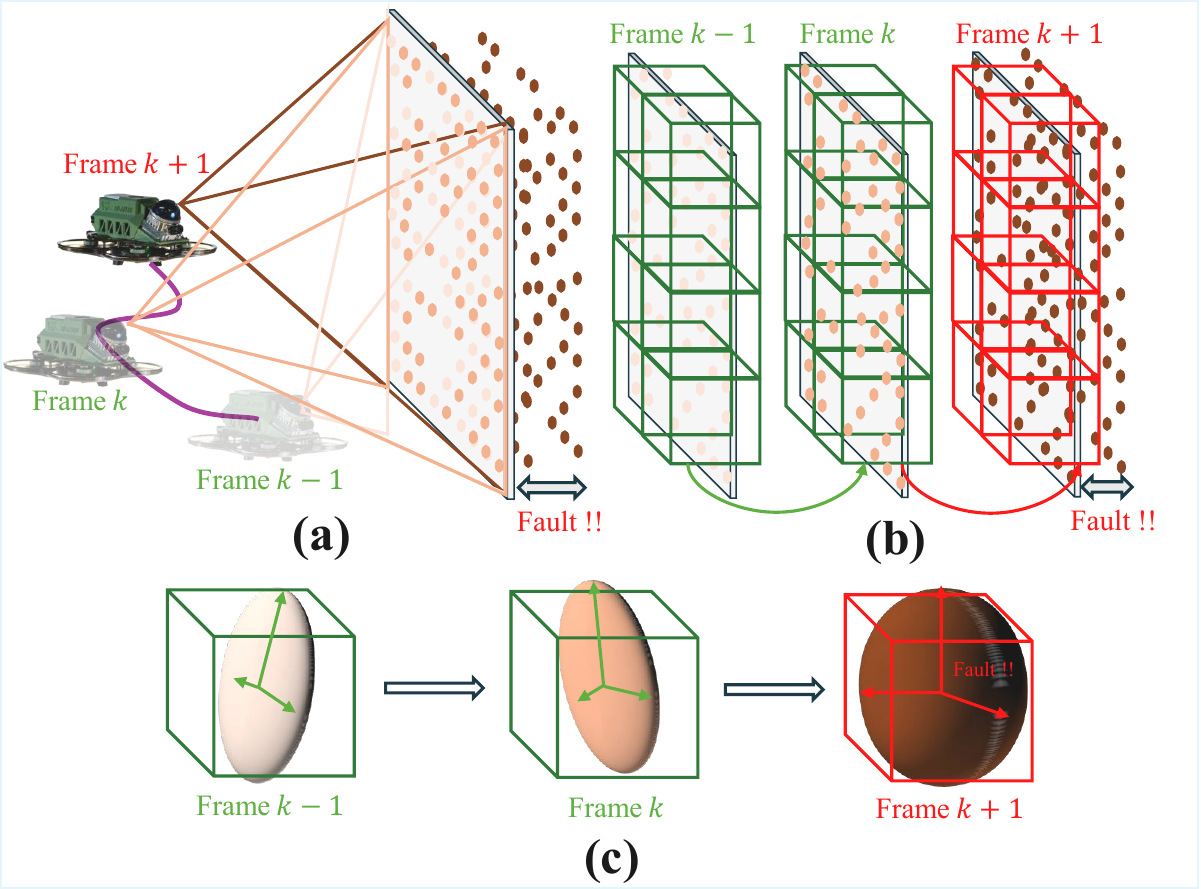}
\caption{Geometric effects of misregistration on a revisit. (a) A pose error at frame $k+1$ displaces a previously observed surface. (b) Voxelization localizes the mismatch. (c) The voxel centroid, orientation, and shape change between nominal (green) and misregistered (red) observations.}
\label{introduction}
\end{figure}

\noindent\textbf{Scene geometry provides a direct signal of SLAM consistency.}
Under nominal operation, repeated observations of stationary structures should remain compatible with the geometry accumulated from earlier visits. When pose errors manifest in the map, surfaces may become displaced, duplicated, or distorted. We therefore formulate failure monitoring as a comparison between current voxel geometry and its historical prediction, without requiring prior maps or estimator-internal variables. The central difficulty is preserving a trustworthy geometric history while allowing the reference to evolve.

\noindent\textbf{Reliable monitoring requires separating evaluation from assimilation.}
Each revisit supplies both evidence for testing and information for refining the map. Updating history before evaluation shifts its centroid toward a displaced scan and increases covariance along the displacement, potentially masking the failure. Freezing history indefinitely instead prevents newly observed surfaces from entering future comparisons. We address this tension by evaluating each revisit against the preceding voxel state and using the resulting evidence to govern subsequent updates. An alarm preserves the historical distributions and their trusted spatial coverage.

\noindent\textbf{We construct a predictive contrast between historical and local geometry.}
A recursive Normal--Inverse-Wishart (NIW) model represents each voxel's centroid, directional spread, and parameter uncertainty. A temporary isotropic model adapts to the current sample's location and overall scale. Their log marginal scores measure support for the same points under historical continuation and a weak local reference. Averaging the temporary-minus-historical contrast over sufficiently supported voxels produces Normalized Drift Evidence (NDE). A positive NDE triggers an alarm at a fixed zero boundary, without per-sequence threshold tuning or a claim of calibrated false-alarm probability.

\noindent\textbf{Contributions.}
The main contributions are summarized as follows:
\begin{enumerate}
\item \textbf{Predictive geometric consistency monitoring.} We develop recursive voxel-level beliefs and a continuation--reference evidence contrast that accounts for uncertainty in historical centroid and directional spread.

\item \textbf{Evidence-gated history maintenance.} We score observations before assimilation and prevent alarmed scans from contaminating voxel distributions or expanding trusted spatial coverage.

\item \textbf{Cross-backend validation and flight recovery.} We evaluate cumulative drift and abrupt failures on self-collected sequences and GEODE across SLAM backends, examine uncertainty and accumulated history through ablations, and demonstrate NDE-triggered odometry isolation and recovery in closed-loop UAV flight.
\end{enumerate}

\section{Related Work}
\label{sec:related_work}

\subsection{Estimator Integrity and Map-Based Localization}

Estimator-based integrity monitoring evaluates how navigation faults affect positioning uncertainty. Particle-filter and Kalman-filter methods propagate candidate measurement faults through the estimation process~\cite{osa1,osa2,osa3}, while solution separation extends fault analysis to filtering and smoothing~\cite{bosdelekidis_solution_separation}. Liu et al.~\cite{tim_liu2026_mhss} consider multiple fault hypotheses to quantify positioning uncertainty in LiDAR simultaneous localization and mapping (SLAM). For stationary scanning systems, an external laser tracker provides independent pose measurements for uncertainty evaluation~\cite{registration_pose_uncertainty}. These approaches offer principled integrity evidence, but depend on estimator-specific uncertainty models or external references.

Prior maps offer an alternative geometric reference. Scan-to-map disagreement can reveal localization failures when a prior map is available~\cite{related_prior1,related_prior2}. Planar structures enable quantitative localization-safety assessment~\cite{prior}, while scan-alignability and scene-dynamics maps identify regions where localization may deteriorate~\cite{castellano_risk_maps}. Li et al.~\cite{us_urban_mapping} combine scan matching with navigation measurements and examine localization against outdated maps along an urban delivery route. These methods require reliable pre-existing geometry.

Online SLAM removes the prior-map requirement but introduces a changing reference. OctoMap recursively integrates range observations into occupancy estimates~\cite{octomap}. Valente et al.~\cite{us_evidential_slam} combine laser and stereo measurements in an evidential grid to distinguish static from dynamic obstacles. Adaptive LiDAR map updates accommodate changing registration conditions~\cite{tim_liu2025_adaptive_map}. When that map also serves as the integrity reference, assimilating a displaced scan prematurely can conceal the inconsistency under examination and contaminate the reference used for subsequent tests.

\subsection{Registration Constraints and Geometric Consistency}

Registration-based indicators assess whether observed geometry adequately constrains motion. Generalized ICP incorporates local surface structure from both point sets~\cite{segal_generalized_icp}. X-ICP identifies weakly constrained directions~\cite{related_match1}; LION uses observability to select odometry sources~\cite{related_match2}; and Hessian-based measures guide degeneracy compensation~\cite{related_hessian1}. Covariance-aware registration and models of point, line, and plane uncertainty account for the reliability of geometric constraints~\cite{tim_wu2025_ualio,tim_zhang2026_geometric_uncertainty}. Cao et al.~\cite{us_cao_laser} introduce laser displacement constraints into visual--inertial estimation. Yet local observability alone need not reveal drift across revisits.

Revisit-based approaches exploit previously observed scene structure. Scan Context recognizes revisited places through the spatial distribution of LiDAR returns~\cite{scan_context}. Reflectivity-based pseudo-images support failure and loop detection in LiDAR--inertial estimation~\cite{tim_dong2023_pve}. Under severe degeneracy, MM-LINS preserves the active map and initiates a new one when conditions improve~\cite{related_hessian2}. These methods support revisit recognition or map protection rather than predictive consistency across observations.

Geometric agreement can also be evaluated after registration. Pairwise misalignment classifiers assess two point clouds~\cite{almqvist_misalignment}, whereas CorAl contrasts local entropy in separate clouds and their union~\cite{adolfsson_coral}. At larger scales, road-surface entropy detects blurred or duplicated structures~\cite{kitsukawa_road_entropy}, and MapEval employs voxel distributions to assess global agreement and local consistency~\cite{hu_mapeval}. In contrast, our monitor evaluates the same current points under the preceding voxel belief and a locally adapted reference. Their predictive evidence difference governs both failure detection and subsequent history updates.

\begin{figure}[!tbp]
\centering
\includegraphics[width=1.0\linewidth]{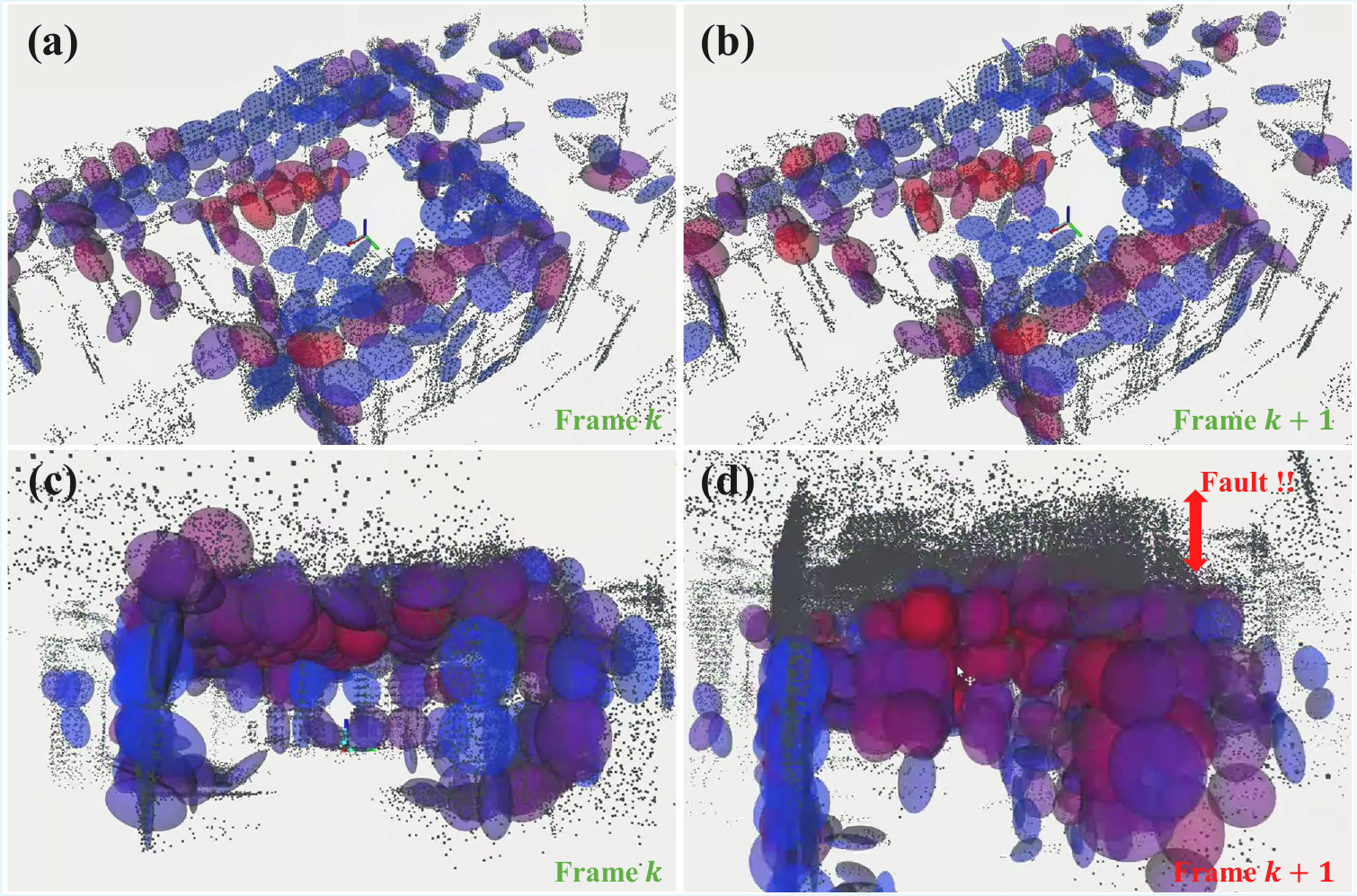}
\caption{Local geometry during nominal mapping and corruption. \textbf{(a)--(b)} Compatible voxel distributions persist while motion extends the observed region. \textbf{(c)--(d)} Corruption changes revisited distributions in centroid, orientation, and thickness. Ellipsoids represent sampled geometry; blue, purple, and red indicate increasing thickness.}
\label{fig:problem_statement}
\end{figure}

\begin{figure*}[!tbp]
\centering
\includegraphics[width=\textwidth]{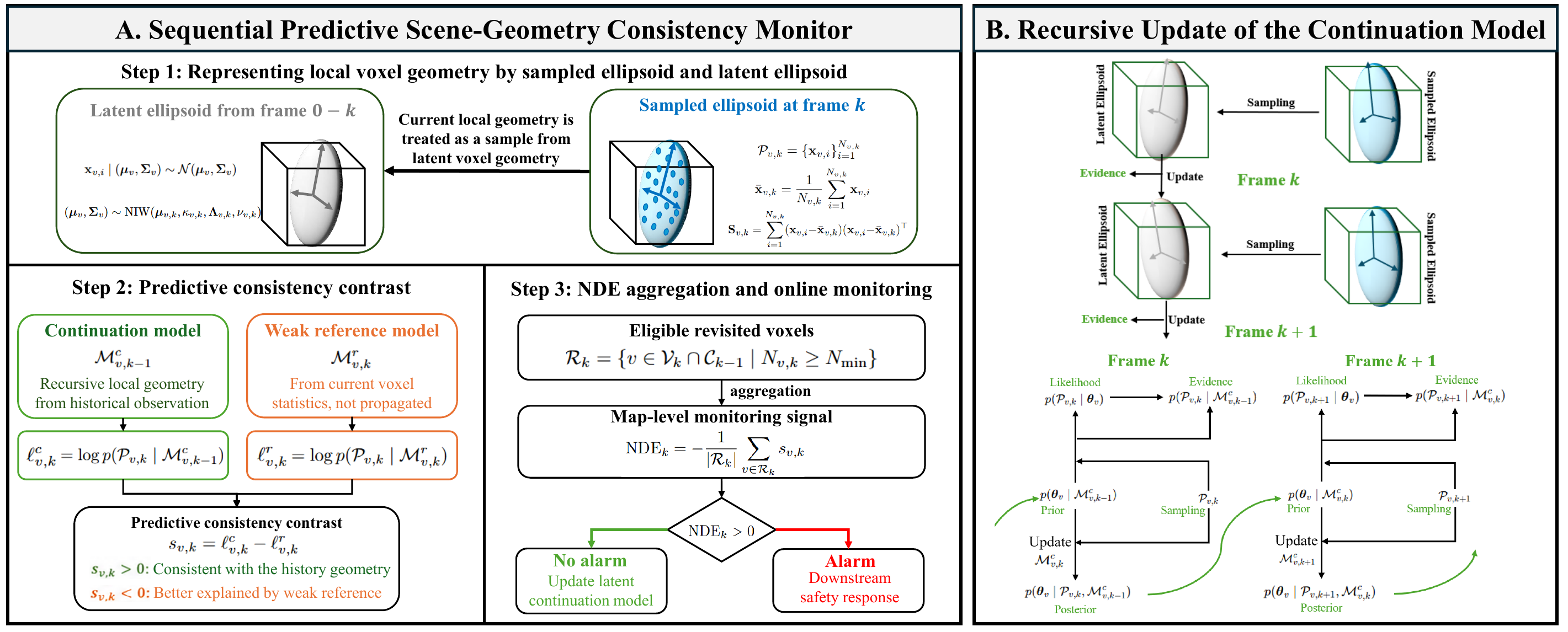}
\caption{Geometric consistency monitoring and recursive updating. \textbf{(A)} Current observations are scored under historical model $\mathcal M^c_{v,k-1}$ and weak, non-propagated reference $\mathcal M^r_{v,k}$. NDE negates the mean contrast $s_{v,k}$ over eligible voxels. Positive NDE triggers an alarm and retains the preceding state. \textbf{(B)} Otherwise, NIW updates incorporate eligible observations for subsequent comparisons.}
\label{fig:method_overview}
\end{figure*}

\subsection{Fault Monitoring and Navigation Recovery}

Monitoring ultimately inform downstream flight control. UAV planners construct feasible trajectories from free-space geometry~\cite{us_uav_trajectory,faster_motion}, making navigation errors consequential for motion execution. Multimodal systems select measurements using subsystem health indicators~\cite{tim_wu2022_multimodal} or reconfigure fusion after faults~\cite{us1}. Deng et al.~\cite{us_deng_lidar} use LiDAR depth clusters to track a UAV during partial GNSS loss. HeRO monitors parallel odometry streams and reinitializes failed estimators while another maintains navigation~\cite{hero_resilience}. Our system instead uses inconsistency in registered scene geometry to trigger odometry isolation and a landing fallback. Successful relocalization against the retained map restores navigation for mission continuation.

\section{Methodology}
\label{sec:methodology}

At each revisit, the monitor evaluates how well the geometry stored from earlier observations explains the current returns. It first scores those returns under two models, then uses their relative support to decide whether to update the history.

\subsection{Monitoring Task and Geometric Input}
\label{sec:problem_statement}
\label{sec:overview}

The input is a sequence of point clouds registered in the fixed map frame of a SLAM backend. The monitor uses the coordinates of revisited surfaces to assess consistency.
It targets duplicated surfaces, persistent offsets, and local deformation (\firstdisplay{Figure}{fig:problem_statement}).

At update $k$, the backend supplies cloud $\mathcal P^{\mathrm{acc}}_k$ accumulated over $T_{\mathrm{acc}}$.
The monitor partitions it on a fixed map-frame grid of side $L_v$.
The defaults are $T_{\mathrm{acc}}=1.0$\,s, $L_v=1.0$\,m, and 1\,Hz; backend preprocessing determines spatial extent.
\firstdisplay{Figure}{fig:method_overview} outlines historical comparison, aggregation, and updating.

Associating voxel distributions at the same map locations across revisits exposes disagreement with stored geometry when historical models remain available and current points provide sufficient support.
Recursive storage extends the reference beyond the accumulation window, preserving evidence of gradual drift.
The window and monitoring cadence affect response timing across transitions to corrupted geometry.

\begingroup
\setlength{\abovedisplayskip}{8.4pt plus 2pt minus 2pt}
\setlength{\belowdisplayskip}{8.4pt plus 2pt minus 2pt}
\setlength{\abovedisplayshortskip}{8.4pt plus 2pt minus 2pt}
\setlength{\belowdisplayshortskip}{8.4pt plus 2pt minus 2pt}
\subsection{Persistent Geometric History}
\label{sec:voxel_geometry}

The historical model represents uncertainty in each voxel's centroid and shape while preserving evidence of drift that remains small between adjacent observations.
For occupied voxel $v\in\mathcal V_k$, the current observation contains $N_{v,k}$ points in $d=3$ dimensions:
\begin{equation}
\mathcal P_{v,k}=\{\mathbf x_{v,i}\}_{i=1}^{N_{v,k}}.
\label{eq:voxel_points}
\end{equation}
The sample centroid is
\begin{equation}
\bar{\mathbf x}_{v,k}
=\frac{1}{N_{v,k}}\sum_{i=1}^{N_{v,k}}\mathbf x_{v,i},
\label{eq:sample_centroid}
\end{equation}
and the scatter matrix records the centered second moments:
\begin{equation}
\mathbf S_{v,k}
=\sum_{i=1}^{N_{v,k}}
(\mathbf x_{v,i}-\bar{\mathbf x}_{v,k})
(\mathbf x_{v,i}-\bar{\mathbf x}_{v,k})^\top.
\label{eq:sample_stats}
\end{equation}
The empirical covariance is $\widehat{\boldsymbol\Sigma}_{v,k}=\mathbf S_{v,k}/N_{v,k}$.
Changes in sampling and viewpoint affect these statistics even when the scene remains unchanged.
We therefore represent local geometry through a latent mean and covariance, $\boldsymbol\theta_v=(\boldsymbol\mu_v,\boldsymbol\Sigma_v)$, with the Gaussian voxel model
\begin{equation}
\mathbf x_{v,i}\mid\boldsymbol\theta_v
\sim\mathcal N(\boldsymbol\mu_v,\boldsymbol\Sigma_v),
\qquad \boldsymbol\Sigma_v\succ0.
\label{eq:latent_gaussian}
\end{equation}
The likelihood assumes conditional independence among the observed points.
Suppressing voxel and update indices, its log form depends only on $(N,\bar{\mathbf x},\mathbf S)$:
\begin{equation}
\begin{aligned}
\log p(\mathcal P\mid\boldsymbol\mu,\boldsymbol\Sigma)
&=-\frac{Nd}{2}\log(2\pi)\\
&\quad-\frac N2\log|\boldsymbol\Sigma|
-\frac12\operatorname{tr}(\boldsymbol\Sigma^{-1}\mathbf S)\\
&\quad-\frac N2(\bar{\mathbf x}-\boldsymbol\mu)^\top
\boldsymbol\Sigma^{-1}(\bar{\mathbf x}-\boldsymbol\mu).
\end{aligned}
\label{eq:voxel_log_likelihood}
\end{equation}

Point count, centroid, and scatter summarize the support, location, and directional dispersion of voxel geometry.
In Equation~\eqref{eq:voxel_log_likelihood}, the covariance-weighted centroid and trace terms evaluate positional and shape disagreement, respectively.

A Normal-Inverse-Wishart (NIW) distribution provides closed-form updates and represents uncertainty in both latent quantities:
\begin{equation}
(\boldsymbol\mu_v,\boldsymbol\Sigma_v)
\sim\mathrm{NIW}(\boldsymbol\mu_{v,k},\kappa_{v,k},
\boldsymbol\Lambda_{v,k},\nu_{v,k}).
\label{eq:niw_model}
\end{equation}
Here, $\boldsymbol\mu_{v,k}$ locates the centroid, $\kappa_{v,k}>0$ controls its concentration, and $(\boldsymbol\Lambda_{v,k},\nu_{v,k})$ parameterize the inverse-Wishart covariance belief.
The joint distribution factorizes as
\begin{equation}
\begin{aligned}
\boldsymbol\Sigma_v
&\sim\mathrm{IW}_d(\boldsymbol\Lambda_{v,k},\nu_{v,k}),\\
\boldsymbol\mu_v\mid\boldsymbol\Sigma_v
&\sim\mathcal N\!\left(\boldsymbol\mu_{v,k},
\frac{\boldsymbol\Sigma_v}{\kappa_{v,k}}\right).
\end{aligned}
\label{eq:niw_factorization}
\end{equation}
We use $\boldsymbol\Lambda_{v,k}\succ0$ and $\nu_{v,k}>d+1$, giving
\begin{equation}
\mathbb E[\boldsymbol\Sigma_v]
=\frac{\boldsymbol\Lambda_{v,k}}{\nu_{v,k}-d-1}.
\label{eq:latent_cov_expectation}
\end{equation}
The ellipsoids in Figure~\ref{fig:method_overview} depict spatial dispersion using this expectation and the centroid mean. The NIW distribution also represents uncertainty about the centroid and covariance themselves. Its concentrations $\kappa_{v,k}$ and $\nu_{v,k}$ grow with the point support incorporated into the history.

For any NIW model with prior parameters $(\boldsymbol\mu_0,\kappa_0,\boldsymbol\Lambda_0,\nu_0)$, let $\boldsymbol\delta=\bar{\mathbf x}-\boldsymbol\mu_0$ denote the current centroid offset.
An observation with statistics $(N,\bar{\mathbf x},\mathbf S)$ gives the standard NIW conjugate updates~\cite{gelman_niw}.
The centroid concentration and covariance degrees of freedom accumulate point support:
\begin{equation}
\kappa_N=\kappa_0+N,
\label{eq:kappa_update}
\end{equation}
\begin{equation}
\nu_N=\nu_0+N.
\label{eq:nu_update}
\end{equation}
The posterior centroid combines the prior location and current sample mean,
\begin{equation}
\boldsymbol\mu_N=\frac{\kappa_0\boldsymbol\mu_0+N\bar{\mathbf x}}{\kappa_0+N},
\label{eq:mu_update}
\end{equation}
while the scale update includes both within-observation scatter and displacement from the prior centroid:
\begin{equation}
\boldsymbol\Lambda_N=\boldsymbol\Lambda_0+\mathbf S
+\frac{\kappa_0N}{\kappa_0+N}
\boldsymbol\delta\boldsymbol\delta^\top.
\label{eq:niw_update}
\end{equation}
These updates explain why assessment precedes assimilation. A displaced surface shifts the posterior centroid in Equation~\eqref{eq:mu_update} toward its new location. Even if its within-scan scatter is unchanged, its centroid offset contributes a positive semidefinite term to the scale matrix in Equation~\eqref{eq:niw_update}. Updating first would therefore alter both parameters used to test the displacement. The monitor computes candidate posteriors but keeps the preceding state as its reference until the alarm decision.

\endgroup
\subsection{Historical and Current Models}
\label{sec:contrast}

For each eligible voxel (Section~\ref{sec:nde}), the monitor scores the current points under the historical and current models.
The continuation model carries the stored history into update $k$:
\begin{equation}
\mathcal M^c_{v,k-1}
=\mathrm{NIW}(\boldsymbol\mu^c_{v,k-1},\kappa^c_{v,k-1},
\boldsymbol\Lambda^c_{v,k-1},\nu^c_{v,k-1}).
\label{eq:continuation_model}
\end{equation}
The superscript $c$ denotes continuation; the model summarizes historical observations incorporated according to the selective updating policy in Section~\ref{sec:online_execution}.
Its score measures compatibility with the current points before their assimilation:
\begin{equation}
\ell^c_{v,k}
=\log p(\mathcal P_{v,k}\mid\mathcal M^c_{v,k-1}).
\label{eq:cont_score}
\end{equation}

The weak reference provides an observation-adaptive baseline for evaluating compatibility with the historical model.
Both marginal scores depend on point count and local geometry; evaluating identical current points removes their common likelihood constant from the contrast.
The reference distribution has the joint density
\begin{equation}
\begin{aligned}
&p(\boldsymbol\mu,\boldsymbol\Sigma\mid\mathcal M^r_{v,k})\\
&=\frac{(\kappa^r_0)^{d/2}|\boldsymbol\Lambda^r_{v,k}|^{\nu^r_0/2}}
{(2\pi)^{d/2}2^{d\nu^r_0/2}\Gamma_d(\nu^r_0/2)}
|\boldsymbol\Sigma|^{-(\nu^r_0+d+2)/2}\\
&\quad\times\exp\!\left[-\frac12\operatorname{tr}
\!\left(\boldsymbol\Lambda^r_{v,k}\boldsymbol\Sigma^{-1}\right)\right]\\
&\quad\times\exp\!\left[-\frac{\kappa^r_0}{2}
(\boldsymbol\mu-\boldsymbol\mu^r_{v,k})^\top
\boldsymbol\Sigma^{-1}(\boldsymbol\mu-\boldsymbol\mu^r_{v,k})\right].
\end{aligned}
\label{eq:reference_niw}
\end{equation}
For positive-definite covariances, the fixed strengths $\kappa^r_0>0$ and $\nu^r_0>d+1$ control the centroid and covariance distributions, respectively.
The reference is centered at the current centroid,
\begin{equation}
\boldsymbol\mu^r_{v,k}=\bar{\mathbf x}_{v,k},
\label{eq:reference_mu}
\end{equation}
and its isotropic scale is set from the mean empirical variance across the three dimensions,
\begin{equation}
\sigma^2_{r,v,k}
=\max\!\left(\epsilon,\frac{\operatorname{tr}(\mathbf S_{v,k})}{dN_{v,k}}\right),
\label{eq:reference_sigma}
\end{equation}
which yields
\begin{equation}
\boldsymbol\Lambda^r_{v,k}
=(\nu^r_0-d-1)\sigma^2_{r,v,k}\mathbf I.
\label{eq:reference_lambda}
\end{equation}
Thus, the expected reference covariance equals $\sigma^2_{r,v,k}\mathbf I$.
The reference inherits the current centroid and total variance. Isotropy makes its expected covariance independent of surface orientation. The historical model, in contrast, carries directional structure from earlier visits into the comparison.
The default values are $\kappa^r_0=10^{-3}$, $\nu^r_0=5$, and $\epsilon=10^{-6}$.

\subsection{Marginal Predictive Scoring}
\label{sec:marginal_evaluation}

Integrating the Gaussian likelihood over each NIW distribution yields the marginal score of the current points.
For either model $H$, the log marginal score is
\begin{equation}
\begin{aligned}
\ell^H&=\log p(\mathcal P\mid H)\\
&=\log\!\int p(\mathcal P\mid\boldsymbol\theta)
p(\boldsymbol\theta\mid H)\,d\boldsymbol\theta.
\end{aligned}
\label{eq:marginal_score_definition}
\end{equation}
Substituting the model parameters into Equations~\eqref{eq:kappa_update}--\eqref{eq:niw_update} gives temporary posterior parameters for evaluating the closed form:
\begin{equation}
\begin{aligned}
\ell^H
&=-\frac{Nd}{2}\log\pi
+\frac d2\log\frac{\kappa^H_0}{\kappa^H_N}\\
&\quad+\frac{\nu^H_0}{2}\log|\boldsymbol\Lambda^H_0|
-\frac{\nu^H_N}{2}\log|\boldsymbol\Lambda^H_N|\\
&\quad+\log\Gamma_d(\nu^H_N/2)
-\log\Gamma_d(\nu^H_0/2),
\end{aligned}
\label{eq:log_marginal}
\end{equation}
where $\Gamma_d(a)=\pi^{d(d-1)/4}\prod_{j=1}^{d}\Gamma(a+(1-j)/2)$ is the multivariate Gamma function.
The two calculations use the same point count, centroid, and scatter. Their initial parameters differ. One model inherits the stored posterior, and the other uses the current centroid and isotropic scale.
The reference score and voxel contrast are
\begin{equation}
\ell^r_{v,k}=\log p(\mathcal P_{v,k}\mid\mathcal M^r_{v,k}),
\label{eq:ref_score}
\end{equation}
\begin{equation}
s_{v,k}=\ell^c_{v,k}-\ell^r_{v,k}.
\label{eq:contrast}
\end{equation}
Positive contrast favors the historical model; negative contrast favors the current reference.
Because the reference uses the scored observation itself, the contrast is data-adaptive and its zero threshold is empirical.
The concentration ratio reflects centroid uncertainty. The scale determinants capture within-scan scatter and centroid displacement through Equation~\eqref{eq:niw_update}.
The Gamma terms normalize for covariance degrees of freedom.

Positive-definite NIW scale matrices ensure well-defined marginal scores for collinear and coplanar samples.
In Equation~\eqref{eq:niw_update}, the scatter and centroid-offset terms are positive semidefinite, so a positive-definite prior scale yields a positive-definite posterior scale.
The regularization floor $\epsilon>0$ preserves positive definiteness of the reference and initialization scales.
The implementation evaluates scale-matrix log determinants through a Cholesky factor:
\begin{equation}
\boldsymbol\Lambda=\mathbf L\mathbf L^\top,
\qquad
\log|\boldsymbol\Lambda|=2\sum_{j=1}^{d}\log L_{jj}.
\label{eq:cholesky_logdet}
\end{equation}
The calculation remains well defined when the empirical covariance is singular.

\begin{figure}[!htb]
    \centering
    \includegraphics[width=\linewidth]{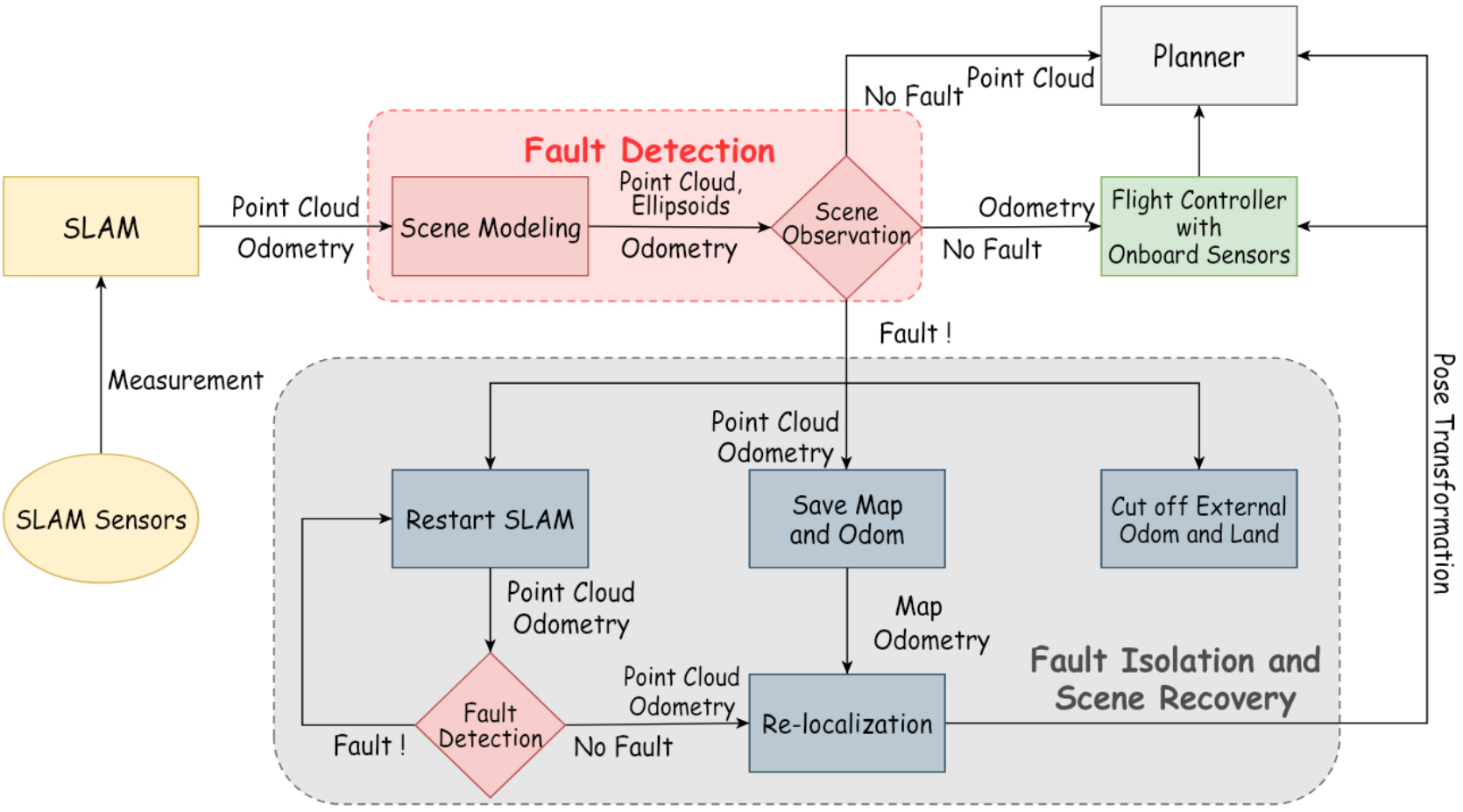}
    \caption{UAV navigation and recovery. An alarm from the geometric monitor triggers odometry isolation and landing using onboard sensing. Relocalization against the retained map restores odometry for takeoff and mission continuation.}
    \label{fig:uav_pipeline}
\end{figure}

\subsection{Aggregation and Geometric Support}
\label{sec:nde}

\begin{algorithm}[!tb]
\caption{Online geometric consistency monitoring}
\label{alg:monitor}
\begin{algorithmic}[1]
\Require Accumulated registered cloud $\mathcal P^{\mathrm{acc}}_k$; stored models $\{\mathcal M^c_{v,k-1}\}_{v\in\mathcal C_{k-1}}$
\State Carry forward $\mathcal C_k\gets\mathcal C_{k-1}$ and all stored models
\State Voxelize $\mathcal P^{\mathrm{acc}}_k$ and compute $(N_{v,k},\bar{\mathbf x}_{v,k},\mathbf S_{v,k})$ for $v\in\mathcal V_k$
\State Form $\mathcal R_k$ using Equation~\eqref{eq:eligible_set}; $L\gets0$
\For{$v\in\mathcal R_k$}
    \State Construct $\mathcal M^r_{v,k}$ using Equations~\eqref{eq:reference_niw}--\eqref{eq:reference_lambda}
    \State $\ell^c_{v,k}\gets\textsc{Score}(\mathcal P_{v,k};\mathcal M^c_{v,k-1})$
    \State $\ell^r_{v,k}\gets\textsc{Score}(\mathcal P_{v,k};\mathcal M^r_{v,k})$
    \State $L\gets L+\ell^c_{v,k}-\ell^r_{v,k}$
\EndFor
\If{$|\mathcal R_k|\ge R_{\min}$}
    \State $\mathrm{NDE}_k\gets-L/|\mathcal R_k|$
    \State $\mathrm{alarm}_k\gets\mathbb I(\mathrm{NDE}_k>0)$
\Else
    \State $\mathrm{NDE}_k\gets\text{undefined}$; $\mathrm{alarm}_k\gets0$
\EndIf
\If{$\mathrm{alarm}_k=0$}
    \For{$v\in\mathcal R_k$}
        \State $\mathcal M^c_{v,k}\gets\textsc{NIW-Update}(\mathcal M^c_{v,k-1},\mathcal P_{v,k})$
    \EndFor
    \For{$v\in\mathcal V_k\setminus\mathcal C_{k-1}$}
        \State $\mathcal M^c_{v,k}\gets\textsc{Initialize}(\mathcal P_{v,k})$
    \EndFor
    \State $\mathcal C_k\gets\mathcal C_{k-1}\cup\mathcal V_k$
\Else
    \State Retain all carried-forward models; issue the alarm to the response module
\EndIf
\State \Return $\mathrm{NDE}_k$, $\mathrm{alarm}_k$, $\{\mathcal M^c_{v,k}\}_{v\in\mathcal C_k}$
\end{algorithmic}
\end{algorithm}

Aggregation evaluates the agreement between current observations and all sufficiently supported historical voxels.
\firstdisplay{Algorithm}{alg:monitor} connects this assessment to the subsequent history-update decision.
Let $\mathcal C_{k-1}$ denote the indices with stored continuation models before update $k$.
The eligible set is
\begin{equation}
\mathcal R_k
=\left\{v\in\mathcal V_k\cap\mathcal C_{k-1}
\;\middle|\;N_{v,k}\ge N_{\min}\right\}.
\label{eq:eligible_set}
\end{equation}
The fixed voxel indices match observations in the same map frame, including revisits after intervening updates.
We use $N_{\min}=10$ points per voxel and require at least $R_{\min}=20$ eligible voxels for a defined map-level score.
All voxels in $\mathcal R_k$ contribute, so the number of aggregated scores varies with the observations.

Eligibility requires historical correspondence and sufficient current point support.
New voxels provide initialization data; supported historical voxels provide comparisons.
Relative overlap measures the proportion revisited, while the absolute eligible count determines whether the aggregate score is defined.

For $|\mathcal R_k|\ge R_{\min}$, Normalized Drift Evidence (NDE) is
\begin{equation}
\mathrm{NDE}_k
=-\frac{1}{|\mathcal R_k|}\sum_{v\in\mathcal R_k}s_{v,k},
\label{eq:nde}
\end{equation}
with the operating rule
\begin{equation}
\mathrm{alarm}_k=\mathbb I(\mathrm{NDE}_k>0).
\label{eq:alarm_rule}
\end{equation}
Positive NDE favors current references over historical models on average.
\firstdisplay{Figure}{fig:uav_pipeline} shows how the UAV uses this alarm.
Its dependence on history, geometry, current support, and reference strengths is evaluated in Section~\ref{subsec:parameter_sensitivity}.

When $|\mathcal R_k|<R_{\min}$, NDE is undefined and the alarm flag is zero. The monitor reports score availability separately. This distinguishes an update with insufficient evidence from a scored revisit that favors the historical model or gives the two models equal support.

\subsection{Scoring and Selective Updating}
\label{sec:online_execution}
\label{sec:algorithm}

The alarm flag governs which candidate posteriors enter the next historical state.
With a zero alarm flag, Equations~\eqref{eq:kappa_update}--\eqref{eq:niw_update} update eligible historical voxels, and newly occupied voxels are initialized.
Undefined NDE permits these operations while retaining its insufficient-support status.
Unobserved or insufficiently supported historical voxels retain their models.
An alarm postpones updates and initialization, preserving the preceding parameters and initialized voxel set for subsequent scoring.

We initialize each newly observed voxel with the current centroid and an isotropic covariance scale:
\begin{equation}
\begin{aligned}
\boldsymbol\mu^c_{v,k}&=\bar{\mathbf x}_{v,k},\\
\boldsymbol\Lambda^c_{v,k}
&=(\nu^c_{\mathrm{init}}-d-1)\sigma^2_{r,v,k}\mathbf I,\\
\kappa^c_{v,k}&=\kappa^c_{\mathrm{init}}=10^{-3},\qquad
\nu^c_{v,k}=\nu^c_{\mathrm{init}}=5.
\end{aligned}
\label{eq:continuation_initialization}
\end{equation}
An initialized voxel becomes eligible at a later update with at least $N_{\min}$ current points.
Algorithm~\ref{alg:monitor} combines scoring and state maintenance.
\textsc{Score} evaluates Equation~\eqref{eq:log_marginal}, \textsc{NIW-Update} applies Equations~\eqref{eq:kappa_update}--\eqref{eq:niw_update}, and \textsc{Initialize} applies Equation~\eqref{eq:continuation_initialization}.

\subsection{Interpreting Geometric Evidence}
\label{sec:aggregate_interpretation}

Persistent environmental changes that alter repeatedly observed structures produce geometric disagreement resembling registration errors.
The NDE response depends on the amount of affected geometry and the strength of local contrasts.
For interpretation, partition eligible voxels into stable-background geometry $\mathcal{B}_k$ and geometry affected by scene changes $\mathcal{D}_k$, including moving or relocated objects:
\begin{equation}
\mathrm{NDE}_k=-\frac{1}{|\mathcal{R}_k|}
\left(\sum_{v\in\mathcal{B}_k}s_{v,k}+\sum_{v\in\mathcal{D}_k}s_{v,k}\right).
\label{eq:background_dynamic_partition}
\end{equation}
A moving object can yield negative local contrasts in the voxels it occupies. The map-level score remains nonpositive when positive contrasts from stable background surfaces offset that contribution.
Nearby or dense moving targets alter this balance when they dominate observations, occlude background, or persistently change accumulated geometry.

\begin{figure*}[t]\centering
	\includegraphics[width=\textwidth]{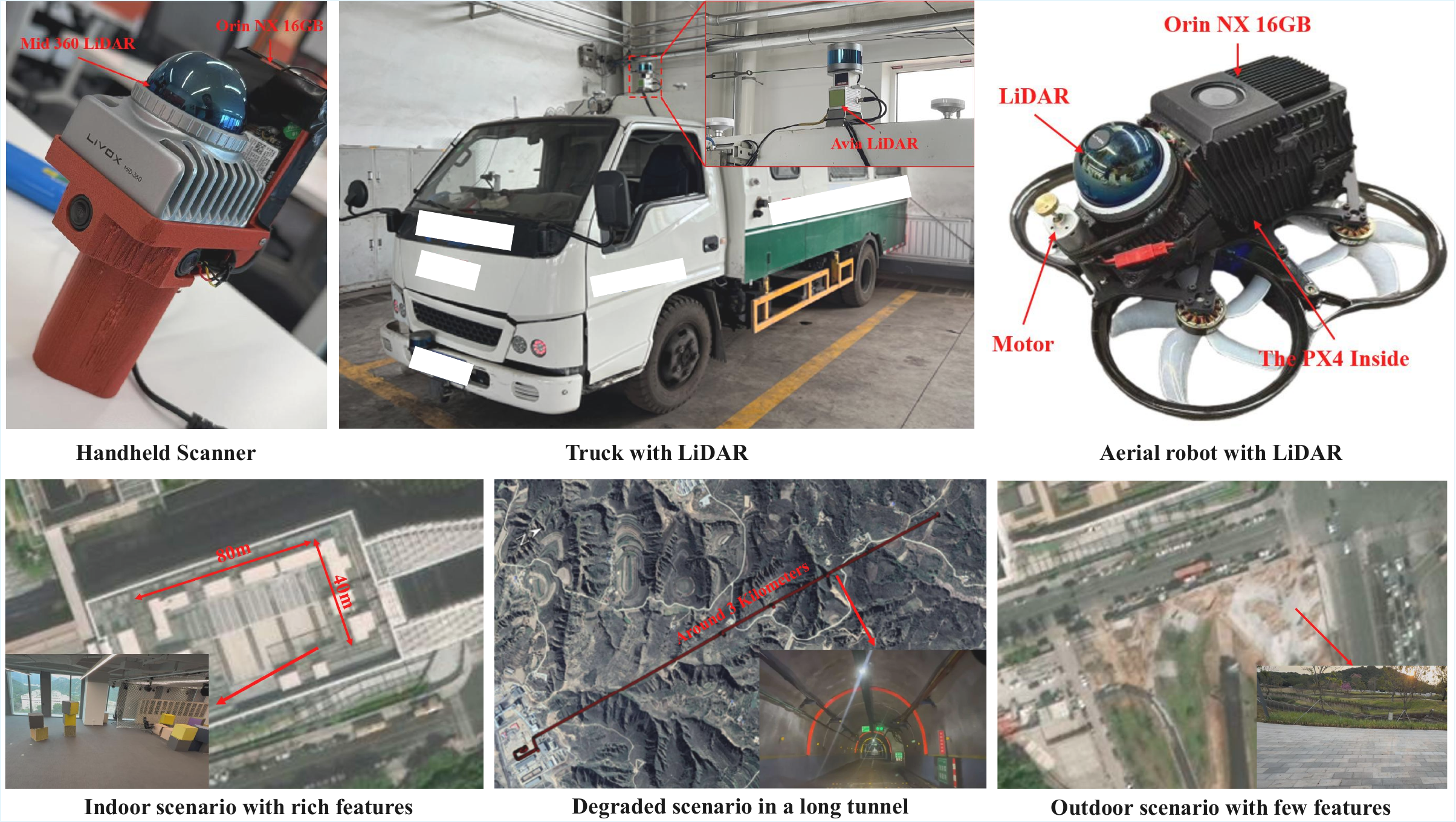}
	\caption{Experimental platforms and environments. Top: handheld, truck-mounted, and aerial LiDAR platforms. Bottom: a feature-rich indoor scene, a long tunnel with weak geometric constraints, and an outdoor scene with limited geometric structure.}
    \label{fig:experiment_setting}
\end{figure*}

\begin{figure*}[t]
\centering
\includegraphics[width=\textwidth]{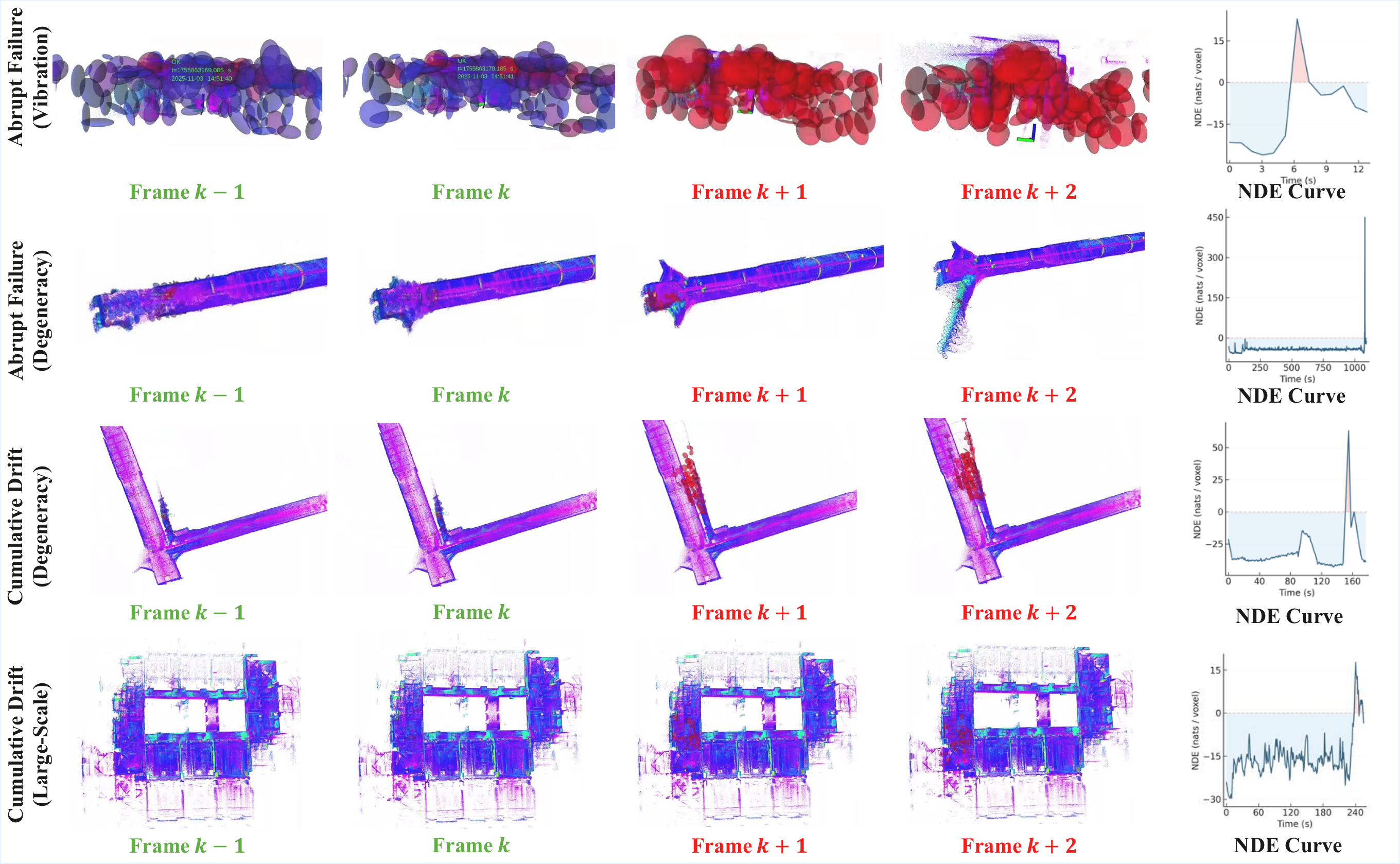}
\caption{Scenarios S1--S4: abrupt corruption under vibration and degeneracy (top two rows), then drift under degeneracy and large-scale traversal (bottom two). Rows pair registered geometry at frames $k-1$ through $k+2$ with NDE. Green/red labels denote observations before/after the reference time; dashed lines mark zero. Blue, purple, and red ellipsoids indicate increasing thickness.}
\label{fig:qualitative_cases}
\end{figure*}

\subsection{Robot Response Interface}
\label{sec:robot_response}

Figure~\ref{fig:uav_pipeline} shows how alarms and score availability inform UAV response under the remaining sensing and control capabilities.
An alarm isolates external FAST-LIO2 odometry from the flight-controller EKF.
The EKF uses onboard sensing for short-term estimation; the controller executes landing using its attitude and altitude control.
SLAM then restarts and relocalizes against the retained pre-failure map, restoring odometry before automatic takeoff and mission continuation.
The retained map supports relocalization, whereas the NIW state preserves historical geometric distributions for subsequent monitoring.

\label{sec:method_end}

\section{Experiments}
\label{sec:evaluation}

\subsection{Platforms and Evaluation Protocol}
\label{subsec:experiment_setting}

The experiments use handheld, truck-mounted, and aerial LiDAR platforms in indoor, tunnel, and outdoor environments (\firstdisplay{Figure}{fig:experiment_setting}).
FAST-LIO2~\cite{fastlio2} processes the self-collected data unless another backend is specified.
The default configuration is $L_v=1.0$\,m, $T_{\mathrm{acc}}=1.0$\,s, a 1\,Hz monitoring rate, $N_{\min}=10$, and $R_{\min}=20$; the decision boundary remains zero throughout.

Reference annotations identify geometric corruption of repeatedly observed static structures, including duplicated surfaces, persistent structural offsets, and deformation.
We establish natural-event references through visual inspection of map snapshots and trajectory replay before examining NDE.
For natural abrupt failures, $t_{\mathrm{ref}}$ marks the first frame with visible geometric corruption.
For cumulative drift, it marks the first frame showing misalignment of revisited static structures that persists in subsequent snapshots; isolated artifacts are excluded.
For injected abrupt failures, $t_{\mathrm{ref}}$ denotes the intervention time, which may precede visible geometric corruption.

For S1--S4, $t_{\mathrm{det}}$ denotes the first NDE alarm in the sequence, and $\Delta t=t_{\mathrm{det}}-t_{\mathrm{ref}}$ reports its offset from the corresponding event reference.
$N_{\mathrm{pre}}$ counts alarms preceding that reference.
For GEODE, labels from visual inspection support sequence-level evaluation of whether an alarm occurs within each run.
Controlled perturbations use logged application times, and UAV response intervals use intervention and response logs.

\begin{figure*}[!tbp]
\centering
\includegraphics[width=.95\textwidth]{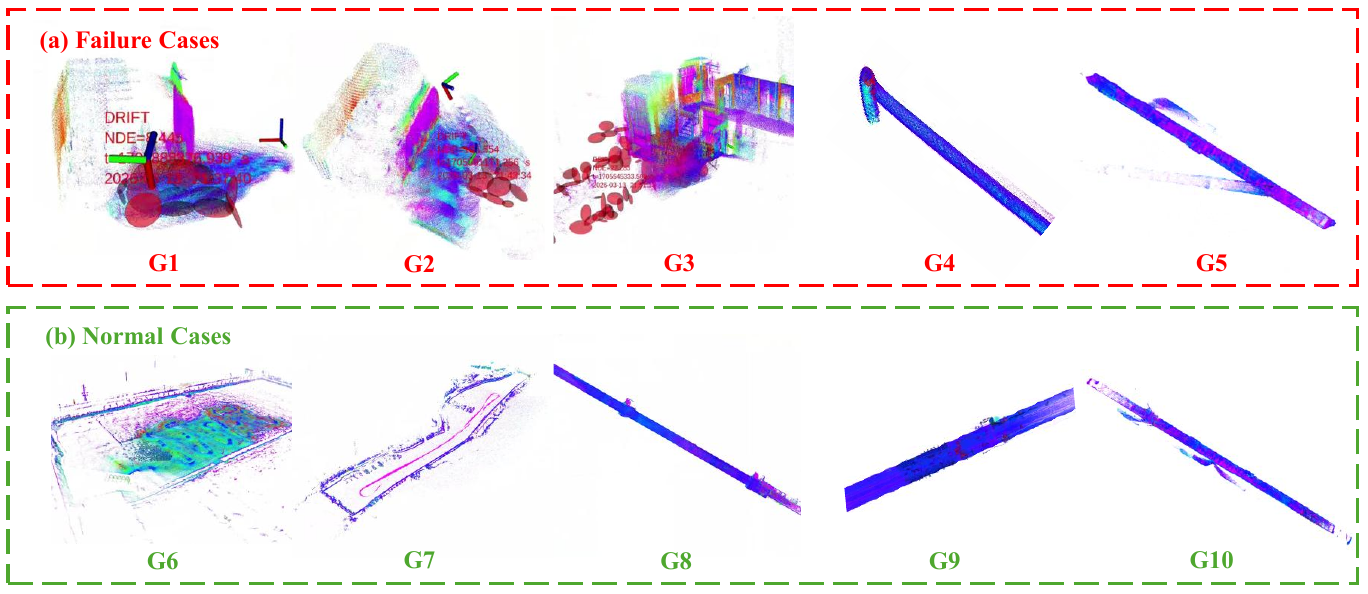}
\caption{Geometry and NDE for ten GEODE runs: \textbf{(a)} corrupted G1--G5 and \textbf{(b)} consistent G6--G10. Map and trajectory inspection determines these groups without using NDE. All runs share the monitoring configuration and zero boundary. Blue, purple, and red ellipsoids indicate increasing thickness.}
\label{fig:benchmark}
\end{figure*}

\subsection{Detection Across Scenes and Backends}
\label{sec:fault_cases}
The 14 real-world sequences evaluated here total approximately 73\,min across four platform types.

\subsubsection{Abrupt Errors and Drift}

Four representative scenes exhibit abrupt misregistration or progressive drift under severe vibration, geometric degeneracy, and large-scale traversal (\firstdisplay{Figure}{fig:qualitative_cases}).
S1 and S2 exhibit abrupt structural changes; S3 and S4 accumulate drift without global loop-closure correction.
In all four scenarios, the first NDE alarm occurs at or after the reference time, with offsets of 0.0, 0.9, 0.2, and 0.4\,s for S1--S4, respectively (\firstdisplay{Table}{tab:self_collected_results}).
In S2, NDE remains below zero for more than 1000\,s before the annotated corruption onset.
The S2 trace shows the distinction between prolonged exposure to weak constraints and the onset of visible surface misalignment.

\begin{figure*}[!tbp]
\centering
\includegraphics[width=\textwidth]{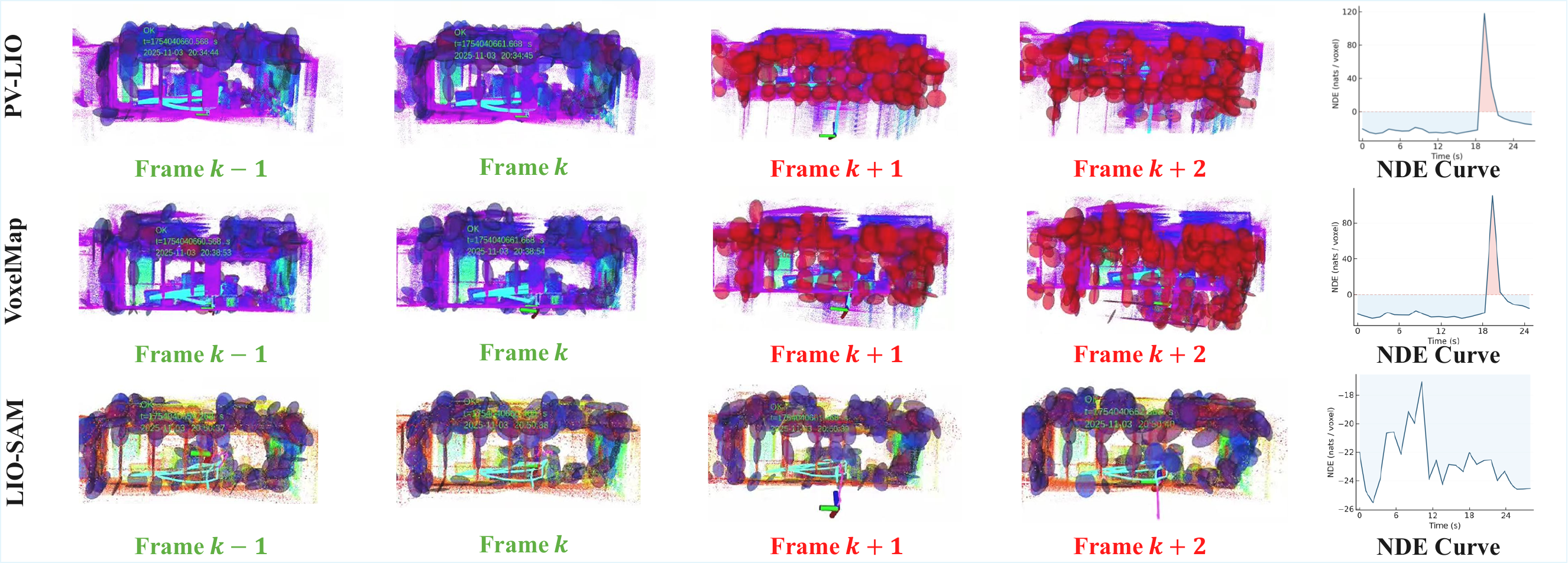}
\caption{Cross-backend comparison on identical input. PV-LIO, VoxelMap, and LIO-SAM rows pair four consecutive registered-map snapshots with NDE under fixed settings. Blue, purple, and red ellipsoids indicate increasing thickness.}
\label{fig:cross_framework_cases}
\end{figure*}

\begin{table*}[!tbp]
\centering
\caption{Event references and NDE responses in the four representative scenes. Times are in seconds. The final two columns give peak evidence and the number of alarms preceding the reference.}
\label{tab:self_collected_results}
\fontsize{8.5}{10.5}\selectfont
\setlength{\tabcolsep}{2pt}
\begin{tabular*}{\linewidth}{@{\extracolsep{\fill}}lllllrrrrrr@{}}
\toprule
ID & Platform & LiDAR & Failure manifestation & Main cause & Duration & $t_{\mathrm{ref}}$ & $t_{\mathrm{det}}$ & $\Delta t$ & $\mathrm{NDE}_{\max}$ & $N_{\mathrm{pre}}$\\
\midrule

S1 & Handheld & MID-360 &
\shortstack[l]{Abrupt\\misregistration} &
\shortstack[l]{Severe\\vibration} &
16.7 & 6.1 & 6.1 & 0.0 & $+18.7$ & 0 \\

S2 & Truck & Livox Avia &
\shortstack[l]{Abrupt\\misregistration} &
\shortstack[l]{Geometric\\degeneracy} &
1107.5 & 1078.3 & 1079.2 & 0.9 & $+450.6$ & 0 \\

S3 & Truck & Livox Avia &
\shortstack[l]{Cumulative\\drift-induced failure} &
\shortstack[l]{Geometric\\degeneracy} &
182.3 & 151.0 & 151.2 & 0.2 & $+63.1$ & 0 \\

S4 & Handheld & MID-360 &
\shortstack[l]{Cumulative\\drift-induced failure} &
\shortstack[l]{Large-scale\\traversal} &
263.3 & 238.3 & 238.7 & 0.4 & $+17.7$ & 0 \\
\bottomrule
\end{tabular*}
\end{table*}

\subsubsection{Public Benchmark and Backend Transfer}
\label{subsec:cross_backend}

Ten GEODE runs~\cite{geode} span six scenarios and three platform types.
Visual inspection before examining NDE identifies five corrupted and five consistent runs.
NDE becomes positive in all five corrupted runs and remains negative throughout all five consistent runs (\firstdisplay{Figure}{fig:benchmark}; \firstdisplay{Table}{tab:geode_results}).
For this sequence-level evaluation, $t_{\mathrm{resp}}$ records the elapsed time from sequence start to the first defined update with positive NDE. For S1--S4 the offsets are measured from the event references.

PV-LIO~\cite{pvlio}, VoxelMap~\cite{voxelmap}, and LIO-SAM~\cite{liosam} process the same input sequence and provide registered clouds to the common monitoring interface.
PV-LIO and VoxelMap produce inconsistent geometry with positive NDE, whereas LIO-SAM produces coherent geometry with negative NDE (\firstdisplay{Figure}{fig:cross_framework_cases}).
The differences in score sign follow the differences in the maps produced from the same sensor data.

\begin{figure}[t]
\centering
\includegraphics[width=1.0\linewidth]{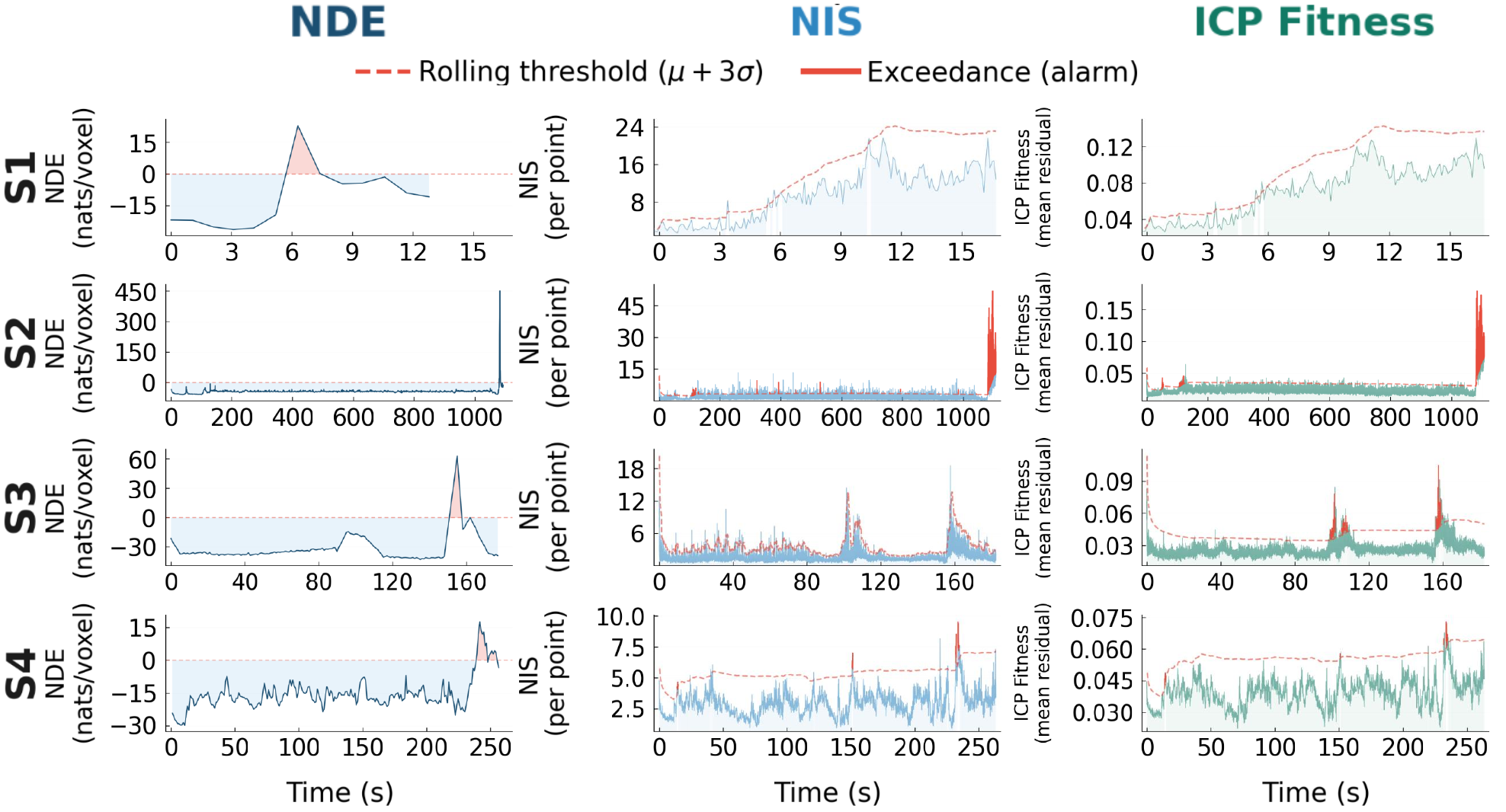}
\caption{NDE and custom residual-indicator responses on S1--S4. NIS-like scores use normalized squared point-to-plane residuals; ICP-fit uses mean absolute residuals. Dashed lines mark the zero NDE boundary and causal rolling thresholds for residual indicators.}
\label{fig:nde_icp}
\end{figure}

\begin{table*}[t]
\centering
\caption{GEODE results under fixed settings. A positive update at any time defines a sequence-level response. The monitor detected 5/5 runs with corrupted geometry and remained alarm-free in 5/5 runs with consistent geometry.}
\label{tab:geode_results}
\fontsize{8.5}{10.5}\selectfont
\setlength{\tabcolsep}{2pt}
\begin{tabular*}{\linewidth}{@{\extracolsep{\fill}}lp{38mm}llr lrrc@{}}
\toprule
ID & Sequence & Platform & Scenario & Duration (s) & Map geometry & $t_{\mathrm{resp}}$ (s) & $\mathrm{NDE}_{\max}$ & Outcome\\
\midrule

G1 & \textit{flat\_\allowbreak{}surfaces\_\allowbreak{}smooth}
& Handheld & Flat surface & 79
& Corrupted & 14.8 & $+20.4$ & TP \\

G2 & \textit{flat\_\allowbreak{}surfaces\_\allowbreak{}aggressive}
& Handheld & Flat surface & 86
& Corrupted & 17.7 & $+105.6$ & TP \\

G3 & \textit{stairs\_\allowbreak{}gamma}
& Handheld & Stairs & 193
& Corrupted & 115.2 & $+112.3$ & TP \\

G4 & \textit{shield5}
& UGV/handheld & Shield tunnel & 35
& Corrupted & 30.7 & $+28.0$ & TP \\

G5 & \textit{Tunneling\_\allowbreak{}tunnel5\_\allowbreak{}gamma}
& UGV & Mine tunnel & 232
& Corrupted & 138.401 & $+785$ & TP \\

G6 & \textit{offroad4}
& UGV & Off-road & 402
& Consistent & -- & $-1.6$ & TN \\

G7 & \textit{inland\_\allowbreak{}waterways\_\allowbreak{}gamma\_\allowbreak{}medium}
& Sailboat & \shortstack[l]{Inland\\waterway} & 836
& Consistent & -- & $-3.7$ & TN \\

G8 & \textit{shield1}
& UGV/handheld & Shield tunnel & 541
& Consistent & -- & $-2.8$ & TN \\

G9 & \textit{Shield\_\allowbreak{}tunnel6\_\allowbreak{}gamma}
& Handheld & Shield tunnel & 224
& Consistent & -- & $-2.1$ & TN \\

G10 & \textit{Tunneling\_\allowbreak{}tunnel1\_\allowbreak{}gamma}
& UGV & Mine tunnel & 208
& Consistent & -- & $-5.9$ & TN \\
\bottomrule
\end{tabular*}
\end{table*}

\subsection{Monitoring Signals and Ablation}
\label{subsec:baseline_monitors}

\subsubsection{Responses of Different Monitors}

\begin{table*}[t]
\centering
\caption{Alarm timing for NDE and four registration indicators on S1--S4. Each scene heading gives duration and reference time in seconds. The window column records an alarm within $t_{\mathrm{ref}}\pm1$\,s. The final block pools alarms and counts scenes with an in-window response.}
\label{tab:comparison}
\fontsize{8.5}{10.5}\selectfont
\setlength{\tabcolsep}{2pt}
\begin{tabular*}{\linewidth}{@{\extracolsep{\fill}}lrrrrcrrrrc@{}}
\toprule
 & \multicolumn{5}{c}{S1 (16.7\,s, 6.1\,s)} & \multicolumn{5}{c}{S2 (1107.5\,s, 1078.3\,s)} \\
\cmidrule(lr){2-6}\cmidrule(l){7-11}
Indicator & $t_{\mathrm{det}}$ (s) & $N_{\mathrm{al}}$ & $N_{\mathrm{pre}}$ & $r_{\mathrm{pre}}$ & Window & $t_{\mathrm{det}}$ (s) & $N_{\mathrm{al}}$ & $N_{\mathrm{pre}}$ & $r_{\mathrm{pre}}$ & Window \\
\midrule
NDE & 6.1 & 1 & \textbf{0} & \textbf{0\%} & Yes & 1079.2 & 1 & \textbf{0} & \textbf{0\%} & Yes \\
NIS-like & 2 & 5 & 3 & 60\% & Yes & 22 & 617 & 589 & 95\% & No \\
ICP-fit & 4 & 3 & 3 & 100\% & Yes & 26 & 184 & 155 & 84\% & Yes \\
X-ICP-inspired~\cite{related_match1} & --- & 0 & 0 & --- & No & 15 & 396 & 368 & 93\% & Yes \\
Hessian-based~\cite{related_hessian1} & --- & 0 & 0 & --- & No & 20 & 285 & 258 & 91\% & Yes \\
\midrule
 & \multicolumn{5}{c}{S3 (182.3\,s, 151.0\,s)} & \multicolumn{5}{c}{S4 (263.3\,s, 238.3\,s)} \\
\cmidrule(lr){2-6}\cmidrule(l){7-11}
Indicator & $t_{\mathrm{det}}$ (s) & $N_{\mathrm{al}}$ & $N_{\mathrm{pre}}$ & $r_{\mathrm{pre}}$ & Window & $t_{\mathrm{det}}$ (s) & $N_{\mathrm{al}}$ & $N_{\mathrm{pre}}$ & $r_{\mathrm{pre}}$ & Window \\
\midrule
NDE & 151.2 & 1 & \textbf{0} & \textbf{0\%} & Yes & 238.7 & 1 & \textbf{0} & \textbf{0\%} & Yes \\
NIS-like & 2 & 94 & 80 & 85\% & Yes & 14 & 17 & 16 & 94\% & No \\
ICP-fit & 26 & 27 & 19 & 70\% & No & 12 & 12 & 12 & 100\% & No \\
X-ICP-inspired~\cite{related_match1} & 5 & 61 & 46 & 75\% & Yes & 32 & 20 & 15 & 75\% & No \\
Hessian-based~\cite{related_hessian1} & 12 & 53 & 38 & 72\% & Yes & --- & 1 & 1 & 100\% & No \\
\bottomrule
\end{tabular*}
\par\smallskip
\begin{tabular*}{\linewidth}{@{\extracolsep{\fill}}lccccc@{}}
\toprule
All scenes & NDE & NIS-like & ICP-fit & X-ICP-inspired & Hessian-based\\
\midrule
$t_{\mathrm{det}}$ (s) & --- & --- & --- & --- & --- \\
$N_{\mathrm{al}}$ & 4 & 733 & 226 & 477 & 339 \\
$N_{\mathrm{pre}}$ & \textbf{0} & 688 & 189 & 429 & 297 \\
$r_{\mathrm{pre}}$ & \textbf{0\%} & 94\% & 84\% & 90\% & 88\% \\
Scenes with an in-window alarm & \textbf{4/4} & 2/4 & 2/4 & 3/4 & 2/4 \\
\bottomrule
\end{tabular*}
\end{table*}

Four custom indicators represent residual monitoring, scan-to-map fitting quality, localizability, and Hessian-based degeneracy.
The NIS-like and ICP-fit indicators use point-to-plane residuals,
\begin{equation}
r_i=\boldsymbol{n}_i^\top(\boldsymbol{x}_i-\boldsymbol{p}_i),
\end{equation}
where $\boldsymbol{x}_i$ is the transformed feature point, $\boldsymbol{p}_i$ the matched map point, and $\boldsymbol{n}_i$ the local plane normal.
Their scores are
\begin{equation}
\mathrm{NIS}_{\mathrm{pp}}=\frac{1}{N}\sum_{i=1}^{N}\frac{r_i^2}{\sigma_i^2},
\qquad
\mathrm{ICP}_{\mathrm{fit}}=\frac{1}{N}\sum_{i=1}^{N}|r_i|.
\end{equation}
The first follows variance-normalized residual monitoring~\cite{osa1,osa2,osa3}; the second measures mean absolute fitting error.
The X-ICP-inspired indicator~\cite{related_match1} raises an alarm for non-localizability along any principal direction.
The Hessian-based indicator~\cite{related_hessian1} raises an alarm for each frame that DBSCAN flags as degenerate, using factors derived from the Hessian condition number. For indicators without intrinsic boundaries, replay uses
\begin{equation}
T(t)=\hat{\mu}_W(t)+3\hat{\sigma}_W(t),
\qquad W=\min(30\tau,0.4T_{\mathrm{total}}),
\end{equation}
where $\tau$ is the output period and rolling statistics use past samples in $[t-W,t]$.
The recorded sequence duration sets the replay window; each indicator uses its specified alarm rule.
Alarm counts use 1\,Hz resolution.
The custom implementations evaluate these monitoring principles within the FAST-LIO2 replay pipeline.
\firstdisplay{Table}{tab:comparison} and \firstdisplay{Figure}{fig:nde_icp} compare temporal responses.
For each indicator, $t_{\mathrm{det}}$ is its first alarm in the sequence, $N_{\mathrm{al}}$ counts all alarms, and $N_{\mathrm{pre}}$ counts those preceding the reference defined in Section~\ref{subsec:experiment_setting}.
The ratio $r_{\mathrm{pre}}=N_{\mathrm{pre}}/N_{\mathrm{al}}$ measures the fraction of alarms preceding the reference.
The Window column records whether an alarm occurred within $t_{\mathrm{ref}}\pm1$\,s.
NDE raises one alarm per sequence, all within the event windows. The fractions preceding the reference are 94\% for NIS-like, 84\% for ICP-fit, 90\% for X-ICP-inspired, and 88\% for Hessian-based monitoring. These indicators responded repeatedly before visible map corruption, particularly during the long approach to the S2 event.

\begin{figure}[!tbp]
\centering
\includegraphics[width=1.0\linewidth]{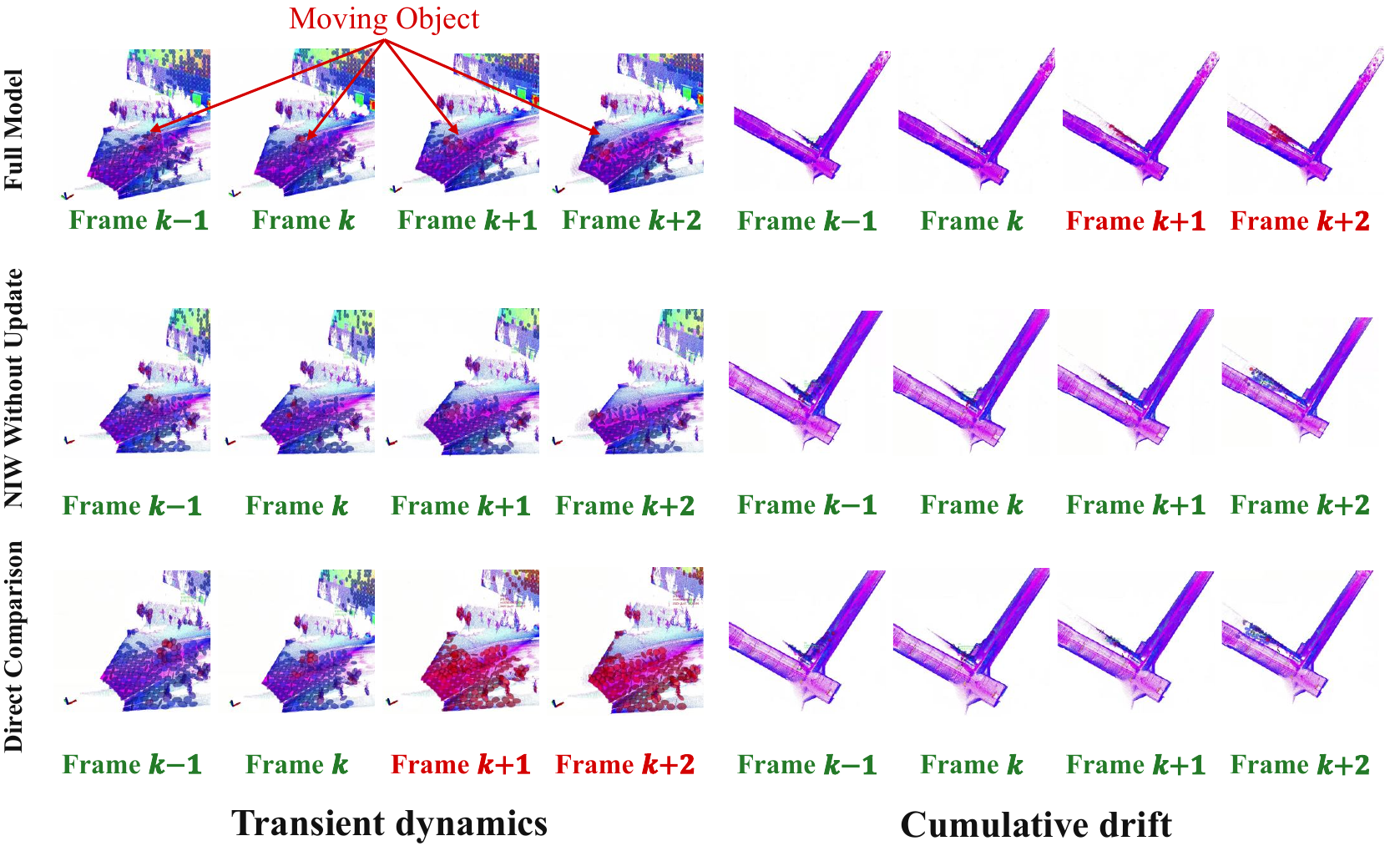}
\caption{Dynamic-scene and cumulative-drift ablation: full NIW with recursive updating, NIW scoring without posterior propagation, and direct comparison of adjacent-frame voxel statistics. Red arrows identify the moving object.}
\label{fig:niw_ablation}
\end{figure}

\subsubsection{Latent Modeling and Recursive History}
\label{subsec:ablation_prior_update}

The \emph{Full Model} combines NIW latent geometry with recursive posterior updating.
\emph{NIW Without Update} retains the predictive score but omits posterior propagation; \emph{Direct Comparison} compares adjacent-frame voxel statistics without latent modeling or recursive history.
In the moving-object example, Direct Comparison raises at least one false alarm and both NIW variants remain alarm-free. Only the Full Model detects accumulated drift (\firstdisplay{Figure}{fig:niw_ablation}).
In the drift example, neighboring observations remain similar. Propagating the posterior preserves earlier surface geometry against which the accumulated displacement becomes apparent.

\subsection{Operating Conditions}
\label{subsec:parameter_sensitivity}

\subsubsection{Controlled Perturbations and Historical Support}
\label{sec:supp_r4_controlled}

\begin{table}[!htbp]
\centering
\caption{Representative controlled pose and support conditions, each with 19 exposure updates.}
\label{tab:controlled_pose_overlap}
\setlength{\tabcolsep}{2.0pt}
\renewcommand{\arraystretch}{1.08}
\footnotesize
\begin{tabular*}{\columnwidth}{@{\extracolsep{\fill}}lccccc@{}}
\toprule
\textbf{Condition} &
$o_{\min}$ &
$E_{\mathrm{med}}$ &
$n_{+}/n_{\mathrm{v}}$ &
$n_{\mathrm{u}}$ &
$\tau_{+}$ \\
& (\%) & & & & (s) \\
\midrule
Unperturbed         & 96.3 & 313 & 0/19  & 0  & --    \\
\midrule
$x=0.10$\,m         & 94.5 & 308 & 0/19  & 0  & --    \\
$x=0.25$\,m         & 87.9 & 294 & 19/19 & 0  & 0.70  \\
$x=10$\,m           & 8.1  & 66  & 18/19 & 0  & 1.70  \\
$x=20$\,m           & 0.0  & 224 & 1/14  & 5  & 16.50 \\
\midrule
Yaw $1^\circ$       & 95.1 & 302 & 0/19  & 0  & --    \\
Yaw $2^\circ$       & 92.5 & 304 & 8/19  & 0  & 4.90  \\
Yaw $5^\circ$       & 83.0 & 277 & 19/19 & 0  & 0.70  \\
\midrule
Retention $q=0.10$  & 88.9 & 40  & 18/19 & 0  & 1.70  \\
Retention $q=0.05$  & 92.9 & 18  & 1/6   & 13 & 1.70  \\
Retention $q=0.02$  & 87.5 & 8   & 0/1   & 18 & --    \\
\bottomrule
\end{tabular*}

\smallskip
\parbox{\columnwidth}{\footnotesize
$o_{\min}$: minimum overlap $E/S$;
$E_{\mathrm{med}}$: median eligible voxels;
$n_{+}/n_{\mathrm{v}}$: positive/defined updates;
$n_{\mathrm{u}}$: undefined updates;
$\tau_{+}$: delay to the first positive update (--: none).
Retention rows include a 0.25\,m translation.
}
\end{table}

Controlled translations, yaw offsets, and support reductions use the first 70\,s of indoor MID-360 data replayed through FAST-LIO2 at $2\times$ speed.
Perturbations are applied only to current registered clouds during $[30,50)$\,s.
Support tests combine a 0.25\,m translation with retention of fraction $q$ of 1\,m voxels selected by spatial hashing (seed 1).
Each condition yields 19 temporally related updates; delays use sensor time from perturbation onset.
Overlap is $o=E/S$, where $S$ counts current voxels with at least 10 points and $E$ counts eligible voxels. NDE is nonpositive for the control, 0.10\,m translation, and $1^\circ$ yaw; 0.25\,m translation and $5^\circ$ yaw produce 19 positive updates, starting at 0.70\,s.
The 20\,m translation produces five undefined updates and one positive update at 16.50\,s among 14 defined updates.
At $q\le0.05$, overlap remains above 87\%, but too few eligible voxels remain for a defined score at most updates.
Scores remain finite for exactly collinear and coplanar points without additional diagonal jitter, and for all 52,267 logged voxel evaluations.
The 21-condition sweep includes a control, nine translations (0.05--20\,m), five yaw offsets ($1^\circ$--$20^\circ$), and six retention settings; \firstdisplay{Table}{tab:controlled_pose_overlap} reports representative conditions.

\begin{figure}[!tbp]
\centering
\includegraphics[width=1.0\linewidth]{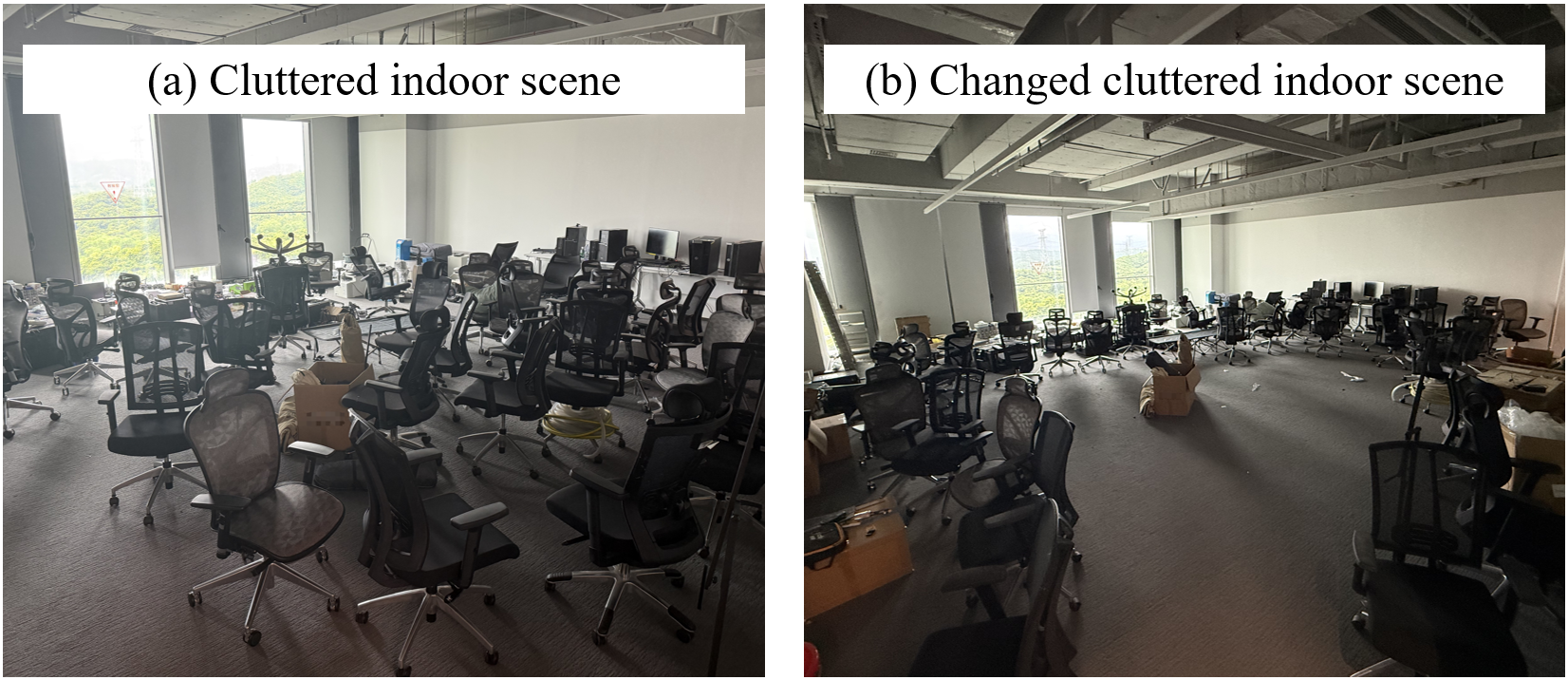}
\caption{Persistent chair rearrangement: \textbf{(a)} original room; \textbf{(b)} rearranged room. The sensor leaves during rearrangement and returns afterward. FAST-LIO2 and the monitor process one continuous recording with unchanged settings.}
\label{fig:supp_chair_rearrangement}
\end{figure}

\subsubsection{Environmental Changes and Moving Traffic}
\label{sec:supp_persistent_environment}

\begin{figure}[!tbp]
\centering
\includegraphics[width=1.0\linewidth]{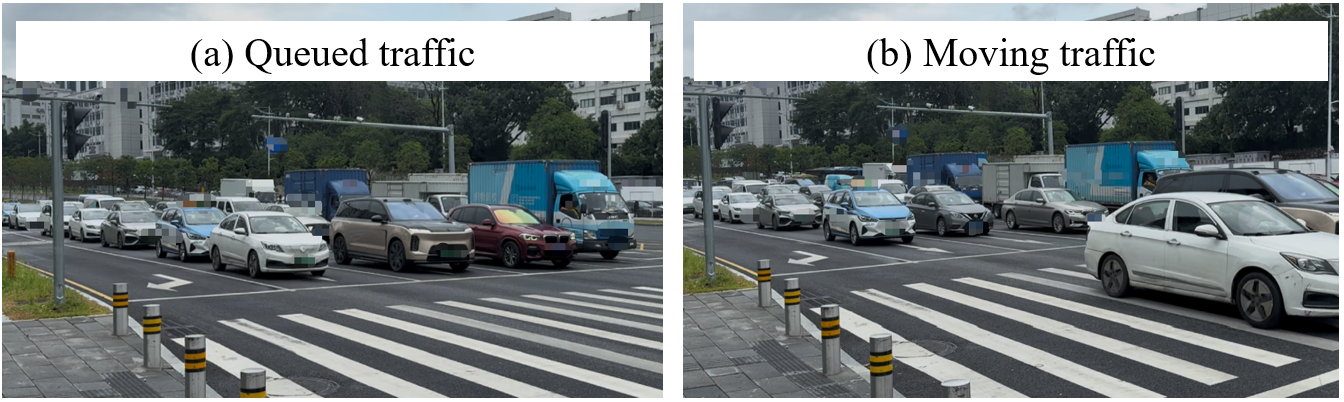}
\caption{Intersection traffic: \textbf{(a)} vehicles queued at a red light; \textbf{(b)} vehicles moving through the junction. Monitoring uses original FAST-LIO2 registered clouds, 1\,m voxels, 1\,s accumulation, and the zero NDE boundary.}
\label{fig:supp_intersection_traffic}
\end{figure}
\begin{table}[!htbp]
\centering
\caption{Responses to chair rearrangement and moving traffic. The complete chair replay also includes the initial and final stationary periods.}
\label{tab:persistent_environment}
\setlength{\tabcolsep}{2.0pt}
\renewcommand{\arraystretch}{1.10}
\footnotesize
\begin{tabular*}{\columnwidth}{@{\extracolsep{\fill}}lcccc@{}}
\toprule
\textbf{Condition} & \textbf{Defined} & \textbf{Positive} & \textbf{NDE} & \textbf{NDE} \\
& /total & /defined & \textbf{median} & \textbf{max} \\
\midrule
Before rearrangement & 46/46 & 0/46 & $-23.813$ & $-15.464$ \\
Transit/intervention & 79/79 & 7/79 & $-27.430$ & $15.445$ \\
After rearrangement & 50/50 & 0/50 & $-22.547$ & $-1.114$ \\
Complete chair replay & 195/197 & 7/195 & $-23.799$ & $15.445$ \\
\midrule
Intersection traffic & 213/-- & 0/213 & $-35.296$ & $-9.582$ \\
\bottomrule
\end{tabular*}
\end{table}

The chair experiment follows a continuous trajectory in a common map frame. The sensor leaves the room during rearrangement and returns afterward (\firstdisplay{Figure}{fig:supp_chair_rearrangement}).
Geometry, trajectory, and IMU evidence define room intervals before NDE analysis: $[10.75,59.25)$\,s before rearrangement and $[142.548537,195.5)$\,s afterward.
NDE remains negative at all 46 updates before rearrangement and all 50 after it (\firstdisplay{Table}{tab:persistent_environment}).
Seven positive updates occur during transit/intervention, while two undefined updates occur during initialization.
In a separate traffic scene, queued vehicles subsequently move through the junction (\firstdisplay{Figure}{fig:supp_intersection_traffic}).
All 213 defined updates have nonpositive NDE under the default configuration.
This response is consistent with stable background evidence offsetting localized traffic changes.

\begin{table}[!htbp]
\centering
\caption{Voxel-size and accumulation-time sensitivity.}
\label{tab:voxel_sensitivity}
\setlength{\tabcolsep}{2.0pt}
\renewcommand{\arraystretch}{1.15}
\footnotesize
\begin{tabular*}{\columnwidth}{@{\extracolsep{\fill}} c c c c c c @{}}
\toprule
& \multicolumn{5}{c}{$T_{\mathrm{acc}}$ (s)} \\
\cmidrule(l){2-6}
$L_v$ (m) & 0.2 & 0.5 & 1.0 & 2.0 & 5.0 \\
\midrule
0.3
& N/D
& N/D
& N/D
& N/D
& N/D \\
0.5
& 7.1 (+1.0)
& 6.9 (+0.8)
& N/D
& N/D
& N/D \\
1.0
& 6.6 (+0.5)
& 6.3 (+0.2)
& \textbf{6.1 (0.0)}
& 6.2 (+0.1)
& N/D \\
2.0
& F/A (1.3)
& 6.3 (+0.2)
& 6.3 (+0.2)
& 6.2 (+0.1)
& F/A (0.0) \\
5.0
& F/A (0.3)
& F/A (0.0)
& F/A (0.0)
& F/A (0.0)
& F/A (0.0) \\
\bottomrule
\end{tabular*}

\par\vspace{2pt}
\begin{minipage}{\columnwidth}
\footnotesize
Entries for valid detections give
$t_{\mathrm{det}}$ ($\Delta t$), both in seconds.
F/A ($t$) denotes a pre-onset first alarm at time $t$;
N/D denotes no detection.
Bold indicates the default setting.
\end{minipage}
\end{table}

\subsubsection{Scale and Reference Parameters}

\firstdisplay{Table}{tab:voxel_sensitivity} evaluates 25 voxel-size and accumulation-time combinations.
Settings with $L_v\in\{1.0,2.0\}$\,m and $T_{\mathrm{acc}}\in\{0.5,1.0,2.0\}$\,s detect the event within 0.0--0.2\,s of the reference.
By contrast, 0.3\,m voxels produce no detection and 5.0\,m voxels produce pre-onset alarms at every tested accumulation time.
Intermediate settings preserve historical support and local geometric resolution.

\begin{table}[!htbp]
\centering
\caption{Sensitivity to weak-reference hyperparameters.}
\label{tab:param_sensitivity}
\setlength{\tabcolsep}{3.0pt}
\renewcommand{\arraystretch}{1.12}
\footnotesize
\begin{tabular*}{\columnwidth}{@{\extracolsep{\fill}} c c c c c @{}}
\toprule
\textbf{Parameter} &
\textbf{Value} &
\boldmath$\mathrm{NDE}_{\min}$ &
\boldmath$\mathrm{NDE}_{\max}$ &
\boldmath$N_{\mathrm{pre}}$ \\
\midrule
\multicolumn{5}{@{}l}{\emph{Scan of $\kappa^r_0$ with $\nu^r_0=5$ fixed}} \\
\midrule
$\kappa^r_0$ & $0.001$ & $-21.4$ & $18.7$ & $0$ \\
$\kappa^r_0$ & $0.005$ & $-18.9$ & $19.5$  & $0$ \\
$\kappa^r_0$ & $0.010$ & $-16.2$ & $22.3$  & $0$ \\
$\kappa^r_0$ & $0.050$ & $-15.8$ & $37.1$  & $0$ \\
$\kappa^r_0$ & $0.100$ & $-17.2$ & $56.9$  & $0$ \\
\midrule
\multicolumn{5}{@{}l}{\emph{Scan of $\nu^r_0$ with $\kappa^r_0=10^{-3}$ fixed}} \\
\midrule
$\nu^r_0$ & $5$   & $-21.4$ & $18.7$  & $0$ \\
$\nu^r_0$ & $10$  & $-21.9$ & $10.3$  & $0$ \\
$\nu^r_0$ & $20$  & $-20.8$ & $15.1$  & $0$ \\
$\nu^r_0$ & $50$  & $-13.4$ & $57.9$  & $\mathbf{2}$ \\
$\nu^r_0$ & $100$ & $5.4$   & $151.2$ & $\mathbf{5}$ \\
\bottomrule
\end{tabular*}
\end{table}

\begin{figure*}[t]
\centering
\includegraphics[width=1\textwidth]{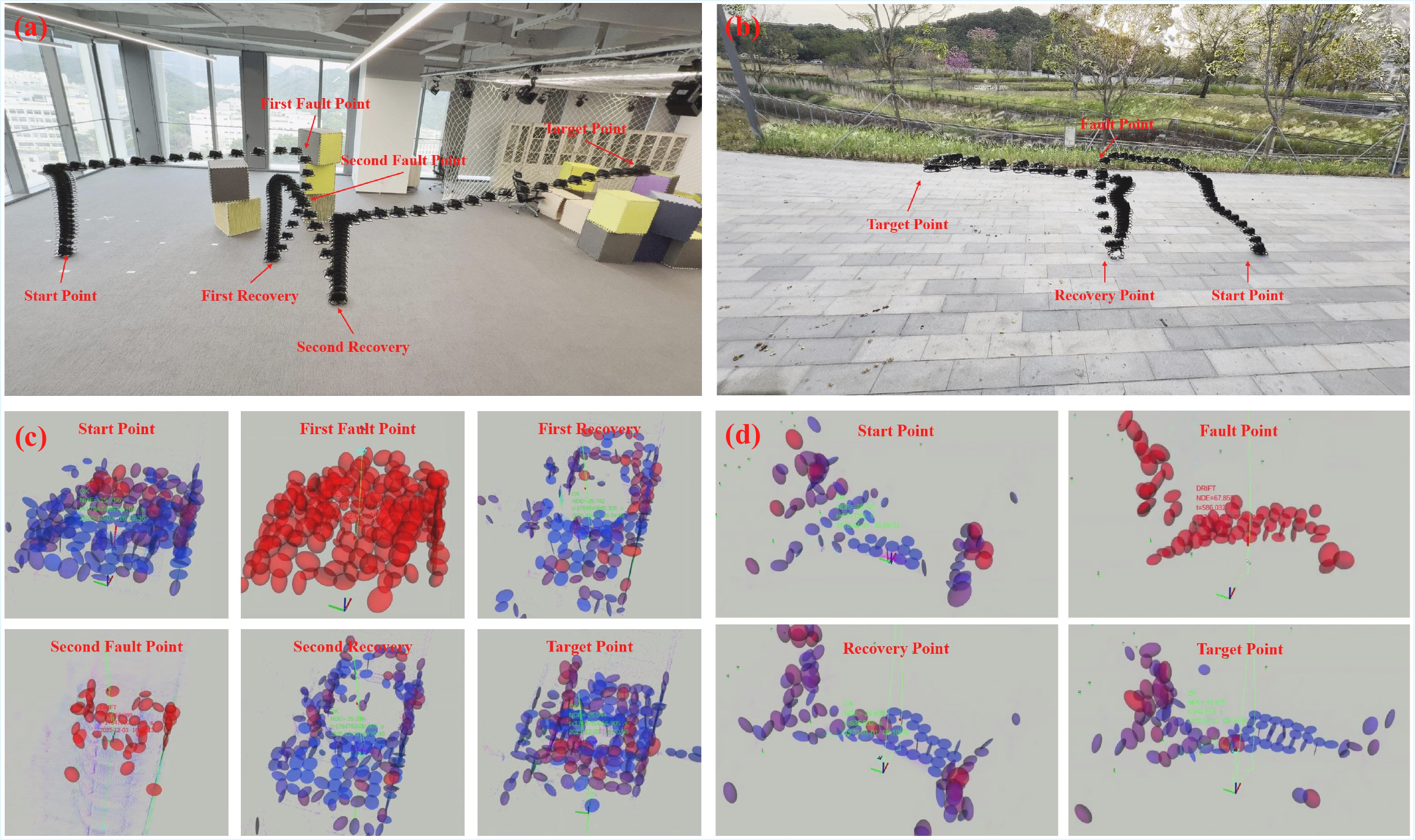}
\caption{Indoor and outdoor UAV missions with vibration-induced SLAM failures. The indoor mission contains two injections. NDE alarms initiate odometry isolation and landing; relocalization against the retained map restores odometry for takeoff and mission continuation.}
\label{fig:uav_application}
\end{figure*}

The weak-reference sweep varies $\kappa^r_0$ with $\nu^r_0=5$ fixed and $\nu^r_0$ with $\kappa^r_0=10^{-3}$ fixed (\firstdisplay{Table}{tab:param_sensitivity}).
Because the reference mean equals the current centroid, $\kappa^r_0$ enters the reference score only through the concentration term in Equation~\eqref{eq:log_marginal}, and larger values raise NDE.
Across the tested range of $\kappa^r_0$, the peak NDE increases from 18.7 to 56.9, the minimum remains between $-21.4$ and $-15.8$, and no pre-onset alarms occur.
Covariance strengths $\nu^r_0\in\{5,10,20\}$ also preserve zero pre-onset alarms, whereas 50 and 100 produce two and five, respectively.
The default $(\kappa^r_0,\nu^r_0)=(10^{-3},5)$ therefore lies within the tested range that avoids pre-onset alarms while maintaining a positive response to the corruption event.

\subsubsection{Complex Local Geometry}
\label{subsubsec:complex_geometry}

\begin{table}[!htbp]
\centering
\caption{Detection and regional evidence in the cluttered room.}
\label{tab:complex_geometry}
\begingroup
\setlength{\tabcolsep}{3pt}
\renewcommand{\arraystretch}{1.05}
\footnotesize
\begin{tabular*}{\columnwidth}{@{\extracolsep{\fill}}lrrrr@{}}
\toprule
\textbf{Input} &
\textbf{Alarms} &
\textbf{Mean NDE} &
\textbf{Furniture} &
\textbf{Wall} \\
\midrule
Unperturbed
& 0/19  & $-23.29$ & $-15.59$ & $-54.95$ \\
0.25\,m translation
& 19/19 & $+60.04$ & $+11.16$  & $+563.01$ \\
$5^\circ$ yaw
& 19/19 & $+71.23$ & $+13.18$ & $+422.98$ \\
\bottomrule
\end{tabular*}
\endgroup
\end{table}

The cluttered-room evaluation compares the central furniture with a fixed wall region.
Three replays of the same 59\,s segment compare unperturbed input, a 0.25\,m body-$x$ translation, and a $5^\circ$ yaw offset during $[30,50)$\,s.
All 19 exposure updates are nonpositive in the control and positive under either offset (\firstdisplay{Table}{tab:complex_geometry}).
The regional diagnostic averages $-s_{v,k}$ over eligible voxels and then nonempty updates.
Regional means indicate stronger inconsistency near walls than around the furniture.
Under both offsets, the furniture region yields positive mean contrasts of $+11.16$ and $+13.18$, more than an order of magnitude below those of the wall region.

\subsection{UAV Validation and Computational Efficiency}
\label{subsec:computation_efficiency}

\subsubsection{UAV Response and Mission Recovery}

\begin{table}[!htbp]
\centering
\caption{Closed-loop flight outcomes indoors and outdoors. Recovery intervals are in seconds and are reported as mean $\pm$ standard deviation.}
\label{tab:uav_response}
\fontsize{8.5}{10.5}\selectfont
\setlength{\tabcolsep}{2pt}
\begin{tabular*}{\linewidth}{@{\extracolsep{\fill}}lrr@{}}
\toprule
 & Indoor & Outdoor\\
\midrule
Trials & 5 & 5 \\
Injected failures per trial & 1 & 1 \\
Trials with an alarm & 5/5 & 5/5 \\
False alarms & 0 & 0 \\
Missions continued & 5/5 & 5/5 \\
\midrule
\multicolumn{3}{@{}l}{\emph{Recovery interval (s)}}\\
Alarm to odometry isolation & $0.31\pm0.06$ & $0.30\pm0.02$ \\
Alarm to landing & $5.50\pm0.23$ & $3.64\pm0.58$ \\
Landing to relocalization & $37.40\pm2.60$ & $49.38\pm4.32$ \\
Relocalization to takeoff & $3.18\pm0.27$ & $4.36\pm0.27$ \\
\bottomrule
\end{tabular*}
\end{table}

The UAV experiments evaluate the odometry-isolation and recovery procedure in Section~\ref{sec:robot_response} using FAST-LIO2.
A motor vibrates the LiDAR assembly at randomly selected times.
The representative indoor mission contains two injected failures followed by recovery and resumed motion; the outdoor mission demonstrates the same response under weaker geometric constraints (\firstdisplay{Figure}{fig:uav_application}). Separately, five indoor and five outdoor trials each contain one injected failure.
ROS and flight-controller logs provide successive response timestamps:
\begin{equation}
t_{\mathrm{alarm}}<t_{\mathrm{iso}}<t_{\mathrm{land}}<t_{\mathrm{reloc}}<t_{\mathrm{takeoff}}.
\end{equation}
\firstdisplay{Table}{tab:uav_response} reports the corresponding intervals.
Every trial completes odometry isolation, landing, relocalization, takeoff, and mission continuation after an alarm.
Across approximately 55 min of flight, no alarm occurs outside the injected failures.
Mean alarm-to-isolation times were 0.31\,s indoors and 0.30\,s outdoors. The longer recovery intervals include landing, SLAM restart, and relocalization.

\begin{table}[!htbp]
\caption{Runtime overhead and timing in representative scenarios. }
\label{tab_efficiency}
\centering
\AtBeginEnvironment{tabular*}{\fontsize{7.5}{9.5}\selectfont}
\setlength{\tabcolsep}{1.2pt}
\renewcommand{\arraystretch}{1.15}

\begin{tabular*}{\columnwidth}{@{\extracolsep{\fill}} l c c c c c c c @{}}
\toprule
& \multicolumn{2}{c}{\textbf{FAST-LIO2}}
& \multicolumn{5}{c}{\textbf{Proposed monitor}} \\
\cmidrule(lr){2-3}\cmidrule(l){4-8}
\textbf{Seq.}
& \shortstack{\textbf{CPU}\\\textbf{(\%)}}
& \shortstack{\textbf{Mem.}\\\textbf{(MB)}}
& \shortstack{\textbf{CPU}\\\textbf{(\%)}}
& \shortstack{\textbf{Mem.}\\\textbf{(MB)}}
& \shortstack{\textbf{Monitoring}\\\textbf{rate (Hz)}}
& \shortstack{\textbf{Processing}\\\textbf{time (ms)}}
& \shortstack{\textbf{Decision}\\\textbf{latency (ms)}} \\
\midrule
S1 & 3--4   & 120--140 & 0.1--0.2 & 15--25 & 1.0 & 4.2 & 7.2 \\
S2 & 12--15 & 130--350 & 0.1--0.2 & 10--20 & 1.0 & 6.1 & 6.5 \\
S3 & 8--12  & 130--155 & 0.1--0.2 & 10--15 & 1.0 & 5.9 & 8.5 \\
S4 & 3--4   & 120--180 & 0.1--0.2 & 15--25 & 1.0 & 4.7 & 6.9 \\
\bottomrule
\end{tabular*}
\end{table}

\subsubsection{Computational Cost}

On an Intel i5-10500, monitoring uses 0.1--0.2\% CPU and 10--25\,MB memory across S1--S4 (\firstdisplay{Table}{tab_efficiency}).
At 1\,Hz, mean processing times are 4.2--6.1\,ms and mean decision latencies are 6.5--8.5\,ms.
Processing covers a complete update; decision latency spans window readiness to NDE/alarm publication, excluding the 1\,s accumulation window.
\section{Discussion and Conclusion}
\label{sec:conclusion}

A geometric history has two roles in online navigation. It supplies the reference for detecting a displaced surface, and it must incorporate new surfaces as the aircraft explores. The proposed monitor reconciles these roles by testing each revisit against the preceding history and using the result to govern assimilation. The comparison scores the same points under the historical NIW model and a temporary reference matched to their centroid and overall spread. An alarm preserves the historical distributions and their spatial coverage for the next assessment.

The experiments connect the two parts of this construction to distinct failure modes. Both NIW variants remained alarm-free in the moving-object example. Only the model that propagated its posterior detected accumulated drift. Across the four event-annotated scenes, the first alarms fell within 0.9\,s of the event references. In closed-loop flight, the aircraft used the alarm to isolate odometry, land, relocalize, and resume its mission.

Two properties of the evidence matter for longer missions. Substantial changes to the scene and errors in registration both displace surfaces relative to the retained history. The flight path also determines how many surfaces are revisited, and successive scores depend on shared history. Spatial coherence and the persistence of local discrepancies offer a basis for distinguishing scene change from drift. A further direction is to calibrate the threshold against false alarms over an entire flight under varying support and temporal dependence.

\section*{Acknowledgments}
This study was supported by the InnoHK initiative of the Innovation and Technology Commission of the Hong Kong Special Administrative Region Government via the Hong Kong Centre for Logistics Robotics, and in part by the Research Grants Council of Hong Kong SAR under Grants 14217922, 14209623, 14200524, and 14209424.

\section*{Conflict of Interest}
The authors declare no other conflicts of interest.

\section*{Data Availability}
The public GEODE sequences are available through the benchmark described by Chen et al.~\cite{geode}. The data collected by the authors are available from the corresponding authors upon reasonable request. The code will be released upon acceptance.

\bibliographystyle{ws-us}
\bibliography{references}

@ARTICLE{fastlio2,
  author={Xu, Wei and Cai, Yixi and He, Dongjiao and Lin, Jiarong and Zhang, Fu},
  journal={IEEE Transactions on Robotics}, 
  title={{FAST-LIO2}: Fast Direct {LiDAR}-Inertial Odometry}, 
  year={2022},
  volume={38},
  number={4},
  pages={2053--2073},
  doi={10.1109/TRO.2022.3141876}}

@ARTICLE{related_prior1,
  author={Akai, Naoki and Morales, Luis Yoichi and Hirayama, Takatsugu and Murase, Hiroshi},
  journal={IEEE Robotics and Automation Letters}, 
  title={Misalignment Recognition Using {Markov} Random Fields With Fully Connected Latent Variables for Detecting Localization Failures}, 
  year={2019},
  volume={4},
  number={4},
  pages={3955--3962},
  doi={10.1109/LRA.2019.2929999}}

@ARTICLE{related_prior2,
  author={Akai, Naoki and Akagi, Yasuhiro and Hirayama, Takatsugu and Morikawa, Takayuki and Murase, Hiroshi},
  journal={IEEE Transactions on Intelligent Transportation Systems}, 
  title={Detection of Localization Failures Using {Markov} Random Fields With Fully Connected Latent Variables for Safe {LiDAR}-Based Automated Driving}, 
  year={2022},
  volume={23},
  number={10},
  pages={17130--17142},
  doi={10.1109/TITS.2022.3164397}}

@ARTICLE{related_hessian1,
  author={Liao, Zongbo and Zhang, Xuanxuan and Zhang, Tianxiang and Li, Zhi and Zheng, Zhenqi and Wen, Zhichao and Li, You},
  journal={IEEE Transactions on Intelligent Transportation Systems}, 
  title={A Real-Time Degeneracy Sensing and Compensation Method for Enhanced {LiDAR} {SLAM}}, 
  year={2025},
  volume={26},
  number={3},
  pages={4202--4213},
  doi={10.1109/TITS.2024.3524394}}

@ARTICLE{related_hessian2,
  author={Ma, Yongxin and Xu, Jie and Yuan, Shenghai and Zhi, Tian and Yu, Wenlu and Zhou, Jun and Xie, Lihua},
  journal={IEEE Transactions on Intelligent Vehicles}, 
  title={{MM-LINS}: A Multi-Map {LiDAR}-Inertial System for Over-Degenerate Environments}, 
  year={2025},
  volume={10},
  number={1},
  pages={472--482},
  doi={10.1109/TIV.2024.3414852}}

@ARTICLE{related_match1,
  author={Tuna, Turcan and Nubert, Julian and Nava, Yoshua and Khattak, Shehryar and Hutter, Marco},
  journal={IEEE Transactions on Robotics}, 
  title={{X-ICP}: Localizability-Aware {LiDAR} Registration for Robust Localization in Extreme Environments}, 
  year={2024},
  volume={40},
  pages={452--471},
  doi={10.1109/TRO.2023.3335691}}

@InProceedings{related_match2,
author="Tagliabue, Andrea and Tordesillas, Jesus and Cai, Xiaoyi and Santamaria-Navarro, Angel and How, Jonathan P. and Carlone, Luca and Agha-mohammadi, Ali-akbar",
title="{LION}: {Lidar}-Inertial Observability-Aware Navigator for Vision-Denied Environments",
booktitle="Experimental Robotics",
year="2021",
publisher="Springer International Publishing",
address="Cham",
pages="380--390",
isbn="978-3-030-71151-1",
doi="10.1007/978-3-030-71151-1_34"
}

@ARTICLE{voxelmap,
  author={Yuan, Chongjian and Xu, Wei and Liu, Xiyuan and Hong, Xiaoping and Zhang, Fu},
  journal={IEEE Robotics and Automation Letters}, 
  title={Efficient and Probabilistic Adaptive Voxel Mapping for Accurate Online {LiDAR} Odometry}, 
  year={2022},
  volume={7},
  number={3},
  pages={8518--8525},
  doi={10.1109/LRA.2022.3187250}}

@misc{pvlio,
  author  = {{HViktorTsoi}},
  title   = {{PV-LIO}},
  year    = {2023},
  howpublished = {\url{https://github.com/HViktorTsoi/PV-LIO}},
  note    = {GitHub repository, accessed 3 November 2025}
}

@INPROCEEDINGS{liosam,
  author={Shan, Tixiao and Englot, Brendan and Meyers, Drew and Wang, Wei and Ratti, Carlo and Rus, Daniela},
  booktitle={2020 {IEEE/RSJ} International Conference on Intelligent Robots and Systems ({IROS})}, 
  title={{LIO-SAM}: Tightly-coupled {Lidar} Inertial Odometry via Smoothing and Mapping}, 
  year={2020},
  pages={5135--5142},
  doi={10.1109/IROS45743.2020.9341176}
}

@ARTICLE{prior,
  author={Zhang, Jiachen and Liu, Xikun and Wen, Weisong and Hsu, Li-Ta},
  journal={IEEE Transactions on Intelligent Vehicles}, 
  title={Safety-Quantifiable Planar-Feature-Based {LiDAR} Localization With a Prior Map for Intelligent Vehicles in Urban Scenarios}, 
  year={2025},
  volume={10},
  number={7},
  pages={3887--3901},
  doi={10.1109/TIV.2024.3467115}}

@ARTICLE{osa1,
  author={Hafez, Osama Abdul and Joerger, Mathieu and Spenko, Matthew},
  journal={IEEE Transactions on Robotics}, 
  title={How Safe Is Particle Filtering-Based Localization for Mobile Robots? An Integrity Monitoring Approach}, 
  year={2024},
  volume={40},
  pages={3372--3387},
  doi={10.1109/TRO.2024.3420798}}

@article{osa2,
  author = {Osama AbdulHafez and Mathieu Joerger and Matthew Spenko},
  title ={Quantifying mobile robot localization safety for an {EKF}-based {SLAM} estimator: An integrity monitoring approach},
  journal = {The International Journal of Robotics Research},
  volume = {44},
  number = {6},
  pages = {972--988},
  year = {2025},
  doi = {10.1177/02783649241287797}
}

@article{osa3,
  author = {Guillermo Duenas Arana and Osama Abdul Hafez and Mathieu Joerger and Matthew Spenko},
  title ={Integrity monitoring for {Kalman} filter-based localization},
  journal = {The International Journal of Robotics Research},
  volume = {39},
  number = {13},
  pages = {1503--1524},
  year = {2020},
  doi = {10.1177/0278364920960517}
}

@article{geode,
author = {Zhiqiang Chen and Yuhua Qi and Dapeng Feng and Xuebin Zhuang and Hongbo Chen and Xiangcheng Hu and Jin Wu and Kelin Peng and Peng Lu},
title ={Heterogeneous {LiDAR} dataset for benchmarking robust localization in diverse degenerate scenarios},
journal = {The International Journal of Robotics Research},
volume = {45},
number = {1},
pages = {6--22},
year = {2026},
doi = {10.1177/02783649251344967}}

@article{us1,
author = {Hachemi, Linda and Guiatni, Mohamed and Nemra, Abedlkrim},
title = {Fault Diagnosis and Reconfiguration for Mobile Robot Localization Based on Multi-Sensors Data Fusion},
journal = {Unmanned Systems},
volume = {10},
number = {01},
pages = {69--91},
year = {2022},
doi = {10.1142/S2301385022500042}
}

@ARTICLE{tim_dong2023_pve,
  author = {Dong, Yanchao and Li, Lingxiao and Liu, Yuhao and Xu, Sixiong and Zuo, Zhelei},
  title = {{PVE-LIOM}: Pseudo-Visual Enhanced {LiDAR}-Inertial Odometry and Mapping},
  journal = {IEEE Transactions on Instrumentation and Measurement},
  year = {2023},
  volume = {72},
  pages = {1--13},
  doi = {10.1109/TIM.2023.3311051}
}

@ARTICLE{tim_wu2022_multimodal,
  author = {Wu, Dongjie and Zhong, Xunyu and Peng, Xiafu and Hu, Huosheng and Liu, Qiang},
  title = {Multimodal Information Fusion for High-Robustness and Low-Drift State Estimation of {UGVs} in Diverse Scenes},
  journal = {IEEE Transactions on Instrumentation and Measurement},
  year = {2022},
  volume = {71},
  pages = {1--15},
  doi = {10.1109/TIM.2022.3205687}
}

@ARTICLE{tim_wu2025_ualio,
  author = {Wu, Qi and Chen, Xieyuanli and Xu, Xiangyu and Zhong, Xinliang and Qu, Xingwei and Xia, Songpengcheng and Liu, Guoqing and Liu, Liu and Yu, Wenxian and Pei, Ling},
  title = {{UA-LIO}: An Uncertainty-Aware {LiDAR}-Inertial Odometry for Autonomous Driving in Urban Environments},
  journal = {IEEE Transactions on Instrumentation and Measurement},
  year = {2025},
  volume = {74},
  pages = {1--12},
  doi = {10.1109/TIM.2025.3544286}
}

@ARTICLE{tim_zhang2026_geometric_uncertainty,
  author = {Zhang, Hongkai and Du, Liang and Hu, Zhengtao and Bao, Sheng and Zhao, Liang and Yuan, Jianjun},
  title = {Tightly Coupled {LiDAR}-Inertial Odometry With Geometric-Uncertainty Modeling},
  journal = {IEEE Transactions on Instrumentation and Measurement},
  year = {2026},
  volume = {75},
  pages = {1--14},
  doi = {10.1109/TIM.2025.3650261}
}

@ARTICLE{tim_liu2025_adaptive_map,
  author = {Liu, Hong and Pan, Shuguo and Wang, Xiang and Gao, Wang and Yu, Baoguo and Zhao, Heng},
  title = {Long-Term Localization Method Integrated With Voxel Mapping {LiDAR} Odometry and Adaptive Updating Map in Diverse Environment},
  journal = {IEEE Transactions on Instrumentation and Measurement},
  year = {2025},
  volume = {74},
  pages = {1--14},
  doi = {10.1109/TIM.2025.3533630}
}

@article{tim_liu2026_mhss,
  author = {Liu, Minzhe and Zhang, Hongjuan and Li, Bijun and Zhao, Zhibo and Qian, Chuang and Liu, Liang},
  title = {Positioning Uncertainty Estimation of {LiDAR} {SLAM} Based on Multifault Hypothesis Solution Separation and Lie Group Theory},
  journal = {IEEE Transactions on Instrumentation and Measurement},
  year = {2026},
  volume = {75},
  pages = {1--15},
  doi = {10.1109/TIM.2026.3655922}
}

@article{almqvist_misalignment,
  author = {Almqvist, H{\aa}kan and Magnusson, Martin and Kucner, Tomasz P. and Lilienthal, Achim J.},
  title = {Learning to detect misaligned point clouds},
  journal = {Journal of Field Robotics},
  year = {2018},
  volume = {35},
  number = {5},
  pages = {662--677},
  doi = {10.1002/rob.21768}
}

@article{castellano_risk_maps,
  author = {Castellano-Quero, Manuel and Kucner, Tomasz Piotr and Magnusson, Martin},
  title = {A {Bayesian} network framework based on {Localization Risk Maps} for anticipating and minimizing localization errors},
  journal = {Robotics and Autonomous Systems},
  year = {2026},
  volume = {206},
  pages = {105707},
  doi = {10.1016/j.robot.2026.105707}
}

@article{bosdelekidis_solution_separation,
  author = {Bosdelekidis, Vasileios and Bryne, Torleiv H. and Sokolova, Nadezda and Johansen, Tor A.},
  title = {Navigation Algorithm-Agnostic Integrity Monitoring based on Solution Separation with Constrained Computation Time and Sensor Noise Overbounding},
  journal = {Journal of Intelligent \& Robotic Systems},
  year = {2022},
  volume = {106},
  pages = {7},
  doi = {10.1007/s10846-022-01692-3}
}

@inproceedings{hero_resilience,
  author = {Santamaria-Navarro, Angel and Thakker, Rohan and Fan, David D. and Morrell, Benjamin and Agha-mohammadi, Ali-akbar},
  title = {Towards Resilient Autonomous Navigation of Drones},
  booktitle = {Robotics Research},
  editor = {Asfour, Tamim and Yoshida, Eiichi and Park, Jaeheung and Christensen, Henrik and Khatib, Oussama},
  series = {Springer Proceedings in Advanced Robotics},
  volume = {20},
  year = {2022},
  publisher = {Springer},
  address = {Cham},
  pages = {922--937},
  doi = {10.1007/978-3-030-95459-8_57}
}

@article{registration_pose_uncertainty,
  author = {Brandst{\"a}tter, Max and Mikschi, Markus and Gabela, Jelena and Linzer, Finn and Neuner, Hans},
  title = {Evaluating Pose Estimation Uncertainty From Automatic Point Cloud Registration in Multi-Sensor Robotic Systems},
  journal = {Journal of Field Robotics},
  year = {2026},
  doi = {10.1002/rob.70360}
}

@inproceedings{segal_generalized_icp,
  author    = {Segal, Aleksandr V. and Haehnel, Dirk and Thrun, Sebastian},
  title     = {Generalized-{ICP}},
  booktitle = {Robotics: Science and Systems V},
  year      = {2009},
  address   = {Seattle, Washington, USA},
  doi       = {10.15607/RSS.2009.V.021}
}

@article{adolfsson_coral,
  author  = {Adolfsson, Daniel and Castellano-Quero, Manuel and Magnusson, Martin and Lilienthal, Achim J. and Andreasson, Henrik},
  title   = {{CorAl}: Introspection for robust radar and lidar perception in diverse environments using differential entropy},
  journal = {Robotics and Autonomous Systems},
  year    = {2022},
  volume  = {155},
  pages   = {104136},
  doi     = {10.1016/j.robot.2022.104136}
}

@article{hu_mapeval,
  author  = {Hu, Xiangcheng and Wu, Jin and Jia, Mingkai and Yan, Hongyu and Jiang, Yi and Jiang, Binqian and Zhang, Wei and He, Wei and Tan, Ping},
  title   = {{MapEval}: Towards Unified, Robust and Efficient {SLAM} Map Evaluation Framework},
  journal = {IEEE Robotics and Automation Letters},
  year    = {2025},
  volume  = {10},
  number  = {5},
  pages   = {4228--4235},
  doi     = {10.1109/LRA.2025.3548441}
}

@article{kitsukawa_road_entropy,
  author  = {Kitsukawa, Yuki and Fukagawa, Minato and Meguro, Junichi and Kato, Shinpei and Edahiro, Masato},
  title   = {Quality Evaluation for {3D} Point Cloud Maps Using Entropy Focused on Road Surfaces},
  journal = {Journal of the Japan Society for Precision Engineering},
  year    = {2025},
  volume  = {91},
  number  = {3},
  pages   = {390--396},
  doi     = {10.2493/jjspe.91.390}
}

@book{gelman_niw,
  author    = {Gelman, Andrew and Carlin, John B. and Stern, Hal S. and Dunson, David B. and Vehtari, Aki and Rubin, Donald B.},
  title     = {Bayesian Data Analysis},
  edition   = {3rd},
  publisher = {Chapman and Hall/CRC},
  address   = {Boca Raton, FL, USA},
  year      = {2013},
  doi       = {10.1201/b16018}
}

@article{us_urban_mapping,
  author = {Li, Qingqing and Pe{\~n}a Queralta, Jorge and Nguyen Gia, Tuan and Zou, Zhuo and Westerlund, Tomi},
  title = {Multi-Sensor Fusion for Navigation and Mapping in Autonomous Vehicles: Accurate Localization in Urban Environments},
  journal = {Unmanned Systems},
  year = {2020}, volume = {8}, number = {3}, pages = {229--237},
  doi = {10.1142/S2301385020500168}
}

@article{us_evidential_slam,
  author = {Valente, Michelle and Joly, Cyril and de La Fortelle, Arnaud},
  title = {Evidential {SLAM} Fusing {2D} Laser Scanner and Stereo Camera},
  journal = {Unmanned Systems},
  year = {2019}, volume = {7}, number = {3}, pages = {149--159},
  doi = {10.1142/S2301385019410012}
}

@article{us_cao_laser,
  author = {Cao, Meng and Zhang, Jia and Chen, Wenjie},
  title = {Visual-Inertial-Laser {SLAM} Based on {ORB-SLAM3}},
  journal = {Unmanned Systems},
  year = {2024}, volume = {12}, number = {5}, pages = {903--912},
  doi = {10.1142/S2301385024500262}
}

@article{us_deng_lidar,
  author = {Deng, Chencheng and Wang, Shoukun and Wang, Junzheng and Xu, Yongkang and Chen, Zhihua},
  title = {{LiDAR} Depth Cluster Active Detection and Localization for a {UAV} with Partial Information Loss in {GNSS}},
  journal = {Unmanned Systems},
  year = {2025}, volume = {13}, number = {2}, pages = {491--503},
  doi = {10.1142/S2301385025500293}
}

@article{us_uav_trajectory,
  author = {Li, Zhaoying and Fei, Yuheng and Shi, Ruoling},
  title = {{UAV} {3D} Trajectory Planning and Optimization Using {A*} Algorithm Based on Free Space Method},
  journal = {Unmanned Systems},
  year = {2024}, volume = {12}, number = {3}, pages = {457--476},
  doi = {10.1142/S2301385024420032}
}

@article{faster_motion,
  author = {Tordesillas, Jesus and Lopez, Brett T. and Everett, Michael and How, Jonathan P.},
  title = {{FASTER}: Fast and Safe Trajectory Planner for Navigation in Unknown Environments},
  journal = {IEEE Transactions on Robotics},
  year = {2022}, volume = {38}, number = {2}, pages = {922--938},
  doi = {10.1109/TRO.2021.3100142}
}

@inproceedings{scan_context,
  author = {Kim, Giseop and Kim, Ayoung},
  title = {Scan Context: Egocentric Spatial Descriptor for Place Recognition Within {3D} Point Cloud Map},
  booktitle = {2018 {IEEE/RSJ} International Conference on Intelligent Robots and Systems ({IROS})},
  year = {2018}, pages = {4802--4809},
  doi = {10.1109/IROS.2018.8593953}
}

@article{octomap,
  author = {Hornung, Armin and Wurm, Kai M. and Bennewitz, Maren and Stachniss, Cyrill and Burgard, Wolfram},
  title = {{OctoMap}: An Efficient Probabilistic {3D} Mapping Framework Based on Octrees},
  journal = {Autonomous Robots},
  year = {2013}, volume = {34}, number = {3}, pages = {189--206},
  doi = {10.1007/s10514-012-9321-0}
}
\balance
\section*{Author Biographies}

\begin{biography}{\includegraphics[width=21mm]{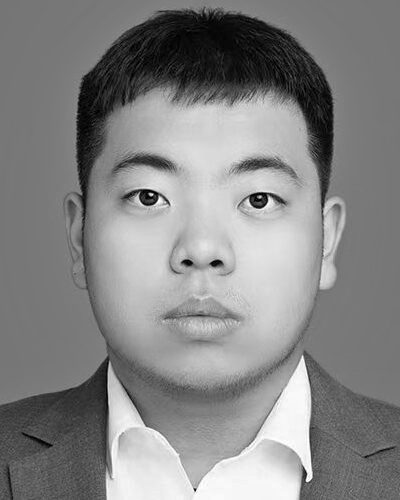}}{\author{Mingqiao Han} received the B.Eng. degree from the University of Nottingham, China, in 2024. He is currently pursuing the Ph.D. degree in mechanical and automation engineering at the Chinese University of Hong Kong (CUHK). His current research interests include SLAM and unmanned systems.}\end{biography}

\begin{biography}{\includegraphics[width=21mm]{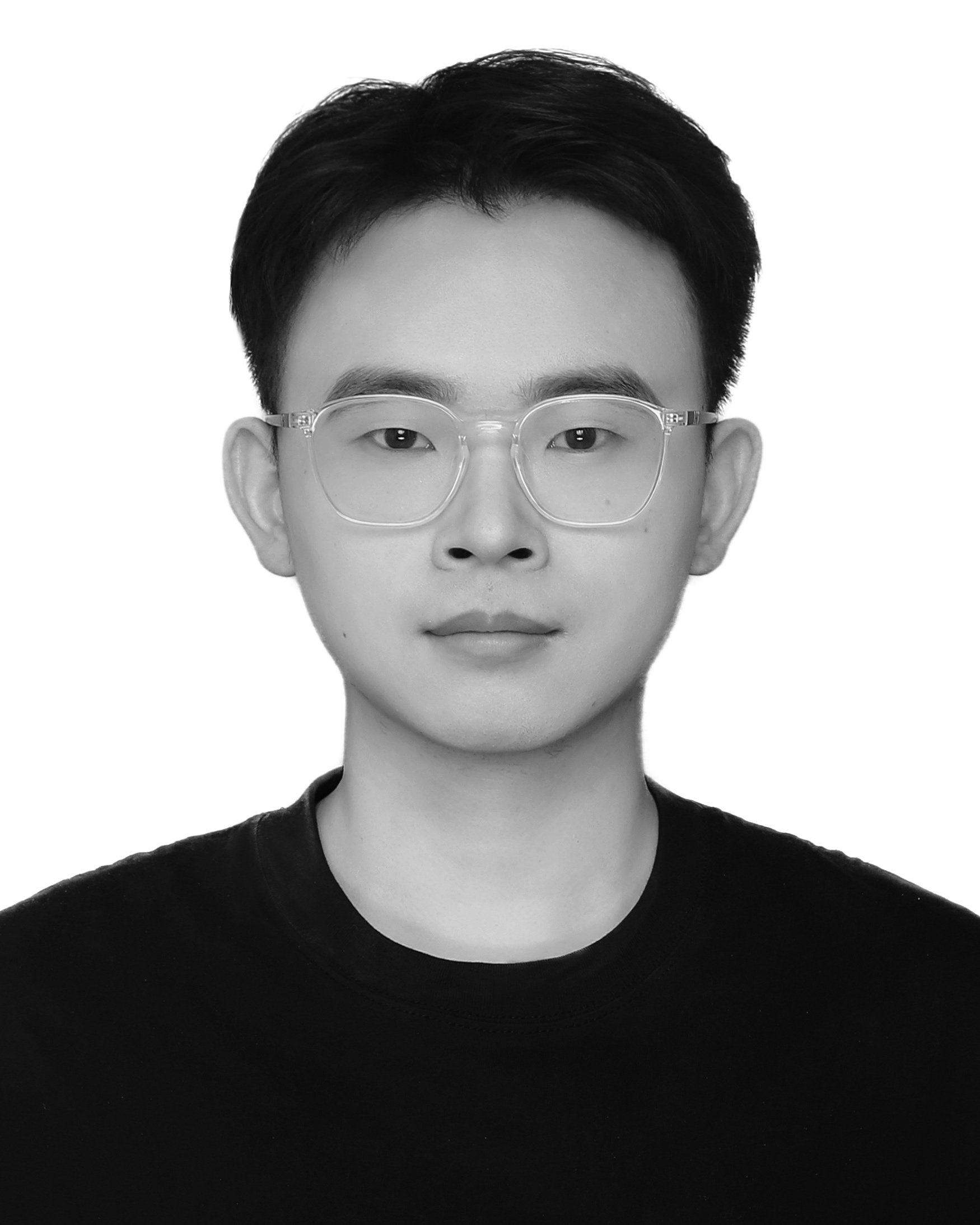}}{\author{Qigeng Duan} received the B.Eng. degree from China University of Geosciences, Beijing, China, in 2019. He worked as a software engineer at NavInfo Co., Ltd., from 2019 to 2021. He is also the founder of Beijing Lishedachuan Co., Ltd. His research interests include SLAM, 3D reconstruction, and digital twins. He is a research assistant in the Unmanned Systems Research Group at the Chinese University of Hong Kong (CUHK).}\end{biography}

\begin{biography}{\includegraphics[width=21mm]{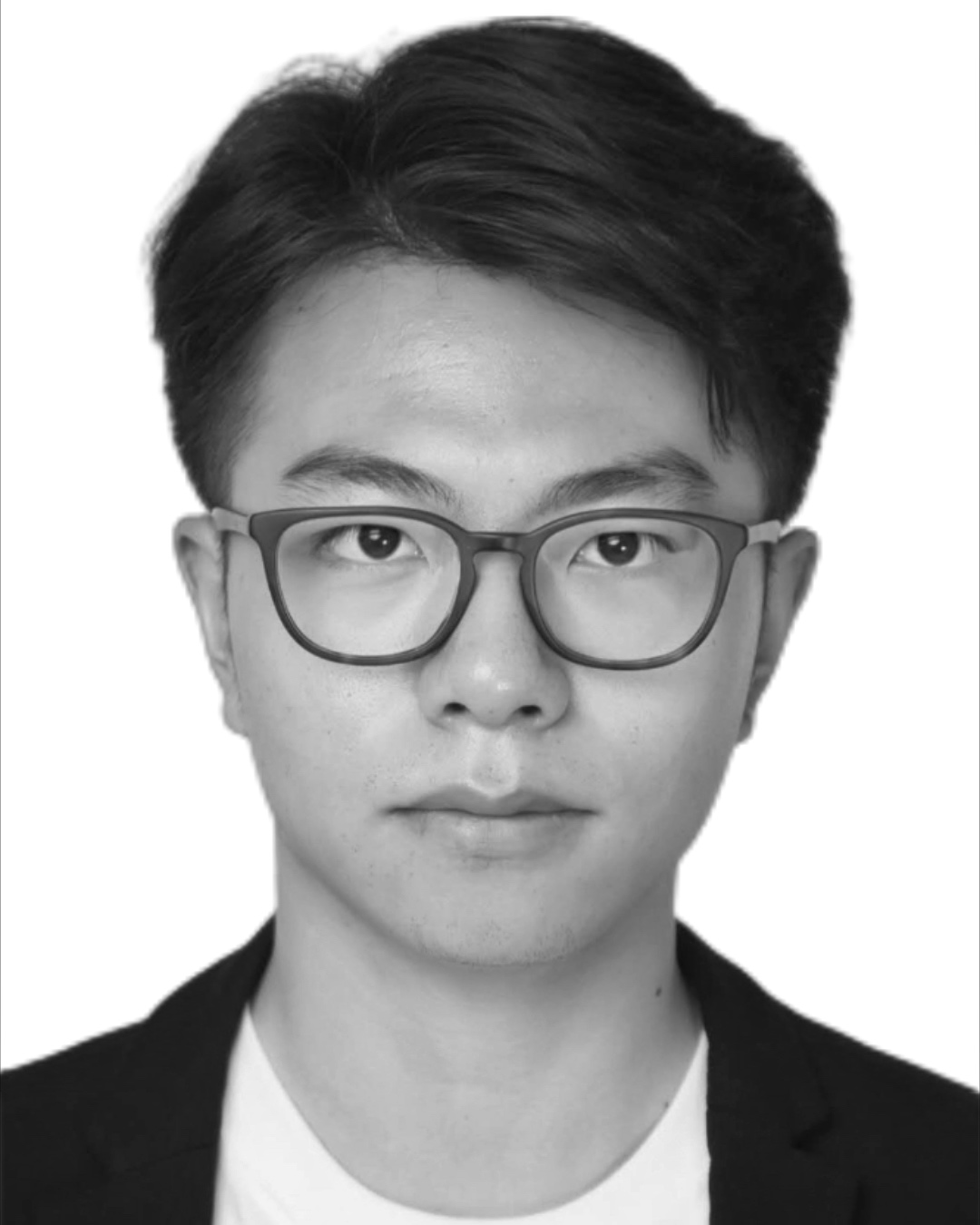}}{\author{Xuchen Liu} is currently a postdoctoral researcher at Pengcheng Laboratory. He received the B.E. degree from Beihang University in 2018 and the M.S. and Ph.D. degrees in mechanical and automation engineering from the Chinese University of Hong Kong (CUHK) in 2020 and 2024, respectively. His research focuses on embodied cognition, autonomous aerial robots, and intelligent multi-robot systems.}\end{biography}

\begin{biography}{\includegraphics[width=21mm]{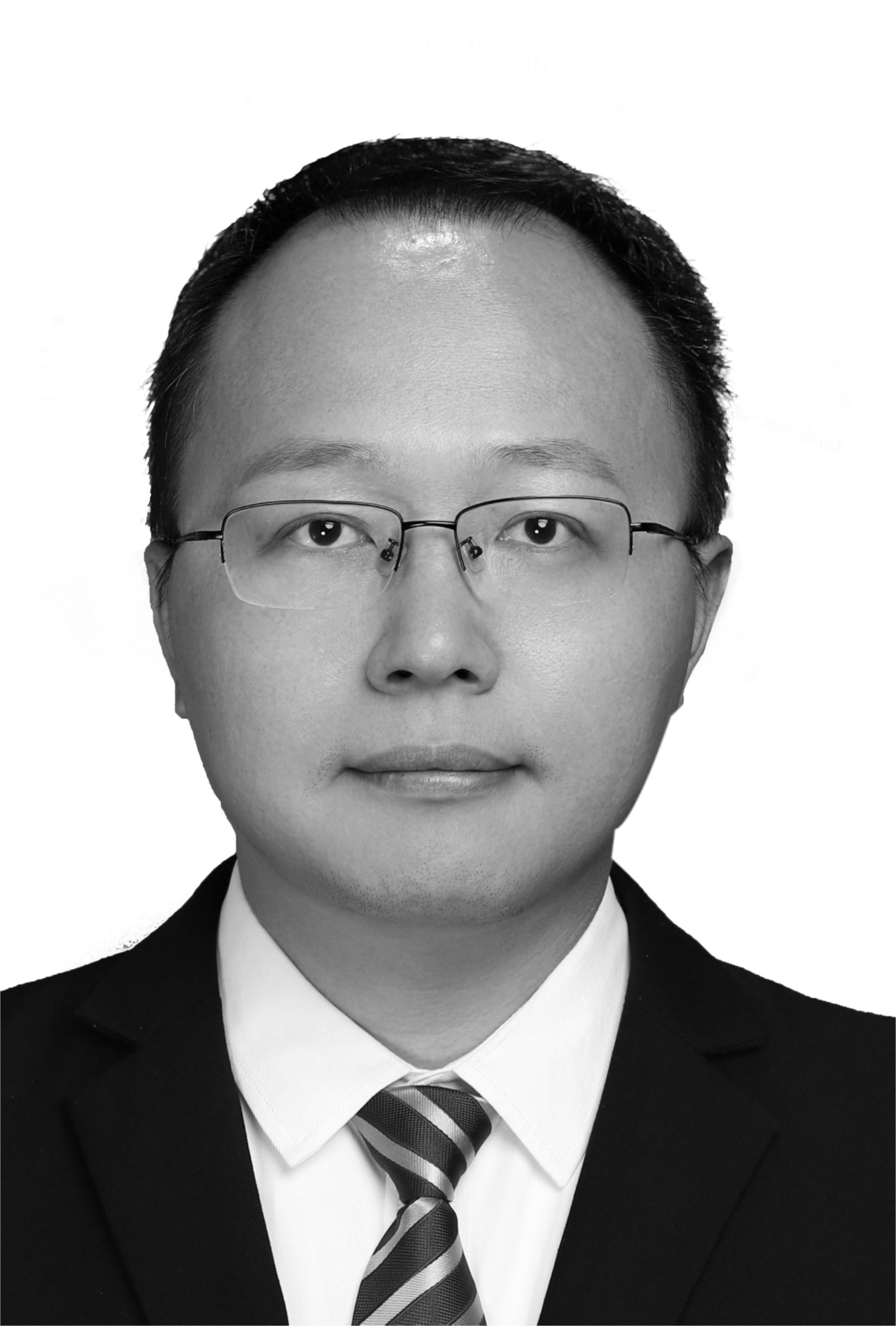}}{\author{Jinqiang Cui} received the B.E. and M.S. degrees from Northwestern Polytechnical University, Xi'an, China, in 2005 and 2008, respectively, and the Ph.D. degree in electrical and computer engineering from the National University of Singapore (NUS) in 2015. He is currently a senior researcher at Pengcheng Laboratory. His current research interests include embodied multi-agent systems, simultaneous localization and mapping, and intelligent multi-robot systems.}\end{biography}

\begin{biography}{\includegraphics[width=21mm]{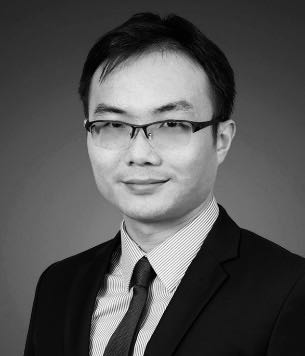}}{\author{Xi Chen} is currently an Assistant Professor of mechanical and automation engineering at the Chinese University of Hong Kong (CUHK), Hong Kong. He has over 10 years of experience in sustainable building technology related to urban energy systems, renewable energy applications in buildings, and built environment modeling. He has led or managed multiple research projects, including ARC, MOST, RGC, and consultancy projects with local government and industry. He has published over 40 papers in peer-reviewed international journals and coauthored a book on green buildings and renewable energy applications.

Dr. Chen has been an Australian Research Council DECRA Fellow and a Fulbright Scholar at Lawrence Berkeley National Laboratory. He also serves on the editorial boards of Buildings, Energies, and Advances in Applied Energy.}\end{biography}

\begin{biography}{\includegraphics[width=21mm]{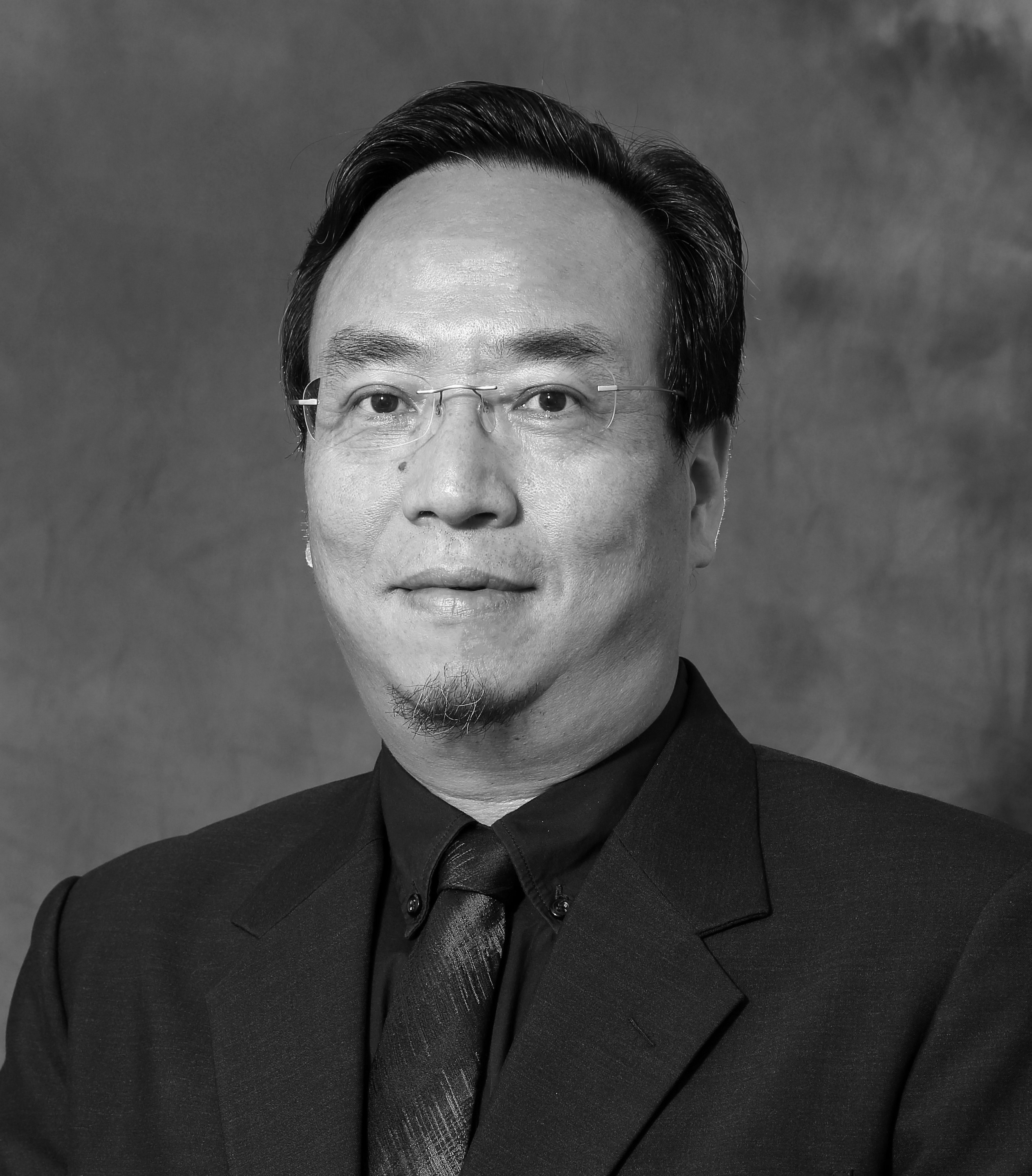}}{\author{Ben M. Chen} (Fellow, IEEE) is currently the Choh-Ming Li Professor of Mechanical and Automation Engineering at the Chinese University of Hong Kong (CUHK), Hong Kong. He was a Provost's Chair Professor with the Department of Electrical and Computer Engineering, National University of Singapore (NUS), before joining CUHK in 2018. He was an Assistant Professor with the Department of Electrical Engineering, State University of New York at Stony Brook, NY, USA, from 1992 to 1993. He has authored or coauthored hundreds of journal and conference articles and a dozen research monographs in control theory and applications, unmanned systems, and financial market modeling. His current research interests are in unmanned systems and their applications.

Dr. Chen is a Fellow of IEEE and a Fellow of the Academy of Engineering, Singapore. He has served on the editorial boards of a dozen international journals, including AUTOMATICA and IEEE TRANSACTIONS ON AUTOMATIC CONTROL. He serves as a Co-Editor-in-Chief of \emph{Unmanned Systems} and an Editor of the International Journal of Robust and Nonlinear Control.}\end{biography}
\end{document}